\documentclass[%
reprint,
amsmath,amssymb,
aip,
apl,
]{revtex4-2}

\usepackage{graphicx}
\usepackage{dcolumn}
\usepackage{bm}
\usepackage[mathlines]{lineno}
\usepackage{xcolor}
\usepackage[normalem]{ulem}

\usepackage{float}
\usepackage{placeins}

\usepackage[utf8]{inputenc}
\usepackage[T1]{fontenc}
\usepackage{mathptmx}
\usepackage{etoolbox}

\newcommand{\alttext}[1]{\par\vspace{2pt}\noindent{\footnotesize\textit{Alt text:} #1\par}}

\begin{document}
\makeatletter
\expandafter\let\csname supp@sv@and\expandafter\endcsname\csname and\endcsname
\expandafter\let\csname supp@sv@affiliation\expandafter\endcsname\csname affiliation\endcsname
\expandafter\let\csname supp@sv@author\expandafter\endcsname\csname author\endcsname
\expandafter\let\csname supp@sv@@AF@join\expandafter\endcsname\csname @AF@join\endcsname
\expandafter\let\csname supp@sv@email\expandafter\endcsname\csname email\endcsname
\expandafter\let\csname supp@sv@@address\expandafter\endcsname\csname @address\endcsname
\expandafter\let\csname supp@sv@maketitle\expandafter\endcsname\csname maketitle\endcsname
\expandafter\let\csname supp@sv@thanks\expandafter\endcsname\csname thanks\endcsname
\makeatother

	\preprint{APS/123-QED}

	\title{Neuromorphic Computing Based on Parametrically-Driven Oscillators and \\ Frequency Combs}

	\author{Mahadev Sunil Kumar}
	\email{mahadevsunilkumar03@gmail.com}
	\homepage{\\https://www.mahadevsunilkumar.com}
	\affiliation{Amrita School of Computing, Amrita Vishwa Vidyapeetham, Kollam, India 690525}

	\author{Adarsh Ganesan}%
	\email{adarsh@dubai.bits-pilani.ac.in}
	\homepage{\\https://www.bits-pilani.ac.in/dr-adarsh-ganesan}
	\affiliation{Department of Electrical and Electronics Engineering, Birla Institute of Technology and Science (BITS) Pilani -- Dubai Campus, Dubai, United Arab Emirates 345055}

	\date{\today}

	\begin{abstract}
	Parametrically driven oscillators exhibit distinct dynamical regimes that strongly influence their computational capability as physical reservoirs. Here, we demonstrate reservoir computing in a two-mode system exhibiting 2:1 parametric resonance using Mackey--Glass, Rössler, and Lorenz chaotic time-series prediction tasks. Mapping the normalized mean squared error over parameter space reveals that the lowest prediction errors consistently occur within the parametric resonance regime, where nonlinear interactions are activated while temporal coherence is preserved. In contrast, although frequency-comb states increase spectral dimensionality, prediction performance degrades in the chaotic comb regime due to loss of phase coherence. We further show that the modulation depth $\Delta \text{F}$, detuning $\kappa$, damping ratio $\gamma_{21}$, and input data rate systematically reshape the accessible dynamical regimes and thereby control the reservoir performance. These results identify parametric resonance as a robust operating regime for oscillator-based neuromorphic computing and establish dynamical regime engineering as a route toward physical reservoir optimization.
	\end{abstract}

	\maketitle

Oscillatory neurocomputing was formalized by Hoppensteadt and Izhikevich \cite{hoppensteadt_oscillatory_1999}, who showed that networks of nonlinear oscillators can function as associative memory systems, where information is encoded in phase relationships and collective synchronization patterns \cite{hoppensteadt_oscillatory_1999, hoppensteadt_synchronization_2001}. Computation in this framework emerges from weakly coupled oscillator dynamics rather than explicit digital operations, enabling pattern recognition through phase locking and frequency interactions. This paradigm has since been implemented across multiple physical platforms, including spintronic oscillators \cite{torrejon_neuromorphic_2017,grollier2020neuromorphic}, coupled VO\textsubscript{2} oscillators \cite{maher2024cmos}, ferroic tunnel junctions \cite{guo2020ferroic}, Belousov-Zhabotinsky chemical oscillators \cite{parrilla2020programmable}, Josephson junctions \cite{chalkiadakis2022dynamical,baxevanis2025inductively}, magnetic skyrmions \cite{azam2018resonate}, feedback field-effect transistor \cite{lee2021fbfet}, transient Joule heating \cite{lashkare2018transient} and microelectromechanical systems (MEMS)-oscillator networks \cite{kumar_autoassociative_2017}.

Despite these advances, the use of parametrically-driven oscillators for neuromorphic computing remains relatively unexplored. Parametric resonance drives the nonlinear energy transfer among modes producing qualitatively distinct dynamical regimes, and has been demonstrated in a remarkably diverse range of physical systems including superconducting circuits \cite{wustmann_parametric_2013}, magnetic systems \cite{chen_parametric_2017}, optics \cite{savchenkov2006parametric,furst2010low}, plasmonics \cite{salandrino2018plasmonic} and MEMS devices \cite{turner_five_1998,ganesan2016observation,ganesan2017excitation}. This phenomenon is also observed in quantum fields \cite{berges_parametric_2003}, the early Universe driving particle production \cite{figueroa_parametric_2017}, dark matter production \cite{dror_parametric_2019}, atmospheric neutrino oscillations \cite{liu_parametric_1998} and in the dynamical models of brain \cite{magazu2024parametric}.

Reservoir computing (RC) is a computational framework in which a fixed high-dimensional nonlinear dynamical system maps input signals into a rich feature space, while only a linear readout layer is trained to generate the desired output \cite{maass_real-time_2002, jaeger_harnessing_2004, lukosevicius_reservoir_2009}. This framework sidesteps the cost of training recurrent network weights and is naturally suited to physical implementation, since any sufficiently nonlinear system with fading memory can serve as the reservoir \cite{tanaka_recent_2019, liang_physical_2024}. Nonlinear oscillators have proven to be particularly effective physical substrates for RC: limit-cycle systems such as the Hopf oscillator support reservoir computation through their intrinsic memory and nonlinearity, with resonance conditions, Arnold tongues, and Farey-sequence frequency ratios all shown to govern computational performance \cite{shougat_dynamic_2022}. More broadly, micromechanical nonlinear resonators \cite{sun_novel_2021, guo_mems_2024}, phononic metastructures \cite{zhang_embodying_2023}, phonon-magnon systems \cite{yaremkevich_-chip_2023}, and spintronic oscillators \cite{torrejon_neuromorphic_2017} have each demonstrated competitive RC performance, establishing nonlinear oscillators as a versatile and hardware-friendly class of physical reservoirs.

Recent work has demonstrated that optical parametric oscillators (OPOs) can serve as ultrafast neuromorphic processors, where cavity feedback and parametric nonlinearities jointly realize recurrent neural network dynamics \cite{parto_ultrafast_2025}. In such systems, sequential input data modulates the pump, and the resulting signal pulses evolve through nonlinear interactions across successive cavity round trips, effectively encoding both memory and nonlinear transformation within the same physical substrate. This architecture enables high-speed processing, with demonstrated operation at GHz rates and sub-nanosecond latencies \cite{parto_ultrafast_2025}. These results highlight the potential of parametric systems as compact, energy-efficient platforms for real-time neuromorphic computation, motivating the exploration of simpler parametrically driven oscillator models and their dynamical regimes for reservoir computing. Parallel to this, optical frequency-comb-based neuromorphic computing has also been recently introduced \cite{shishavan_optical_2025}. These frequency combs provide a high-dimensional set of coherent spectral modes that can serve as computational degrees of freedom. In particular, they can act as delay-free reservoirs, where intrinsic cavity dynamics provides both memory and nonlinear feature expansion for tasks such as chaotic time-series prediction. Also, the nature of comb regimes: coherent versus chaotic, plays a critical role in determining the performance of neuromorphic computing systems.

In this work, we investigate how different dynamical regimes of parametrically-driven oscillators support neuromorphic computation. We model a two-mode system exhibiting 2:1 parametric resonance and use it as a physical reservoir. We systematically vary the drive parameters across the Hopf bifurcation boundary to explore computation in the sub-threshold, parametric resonance, and frequency-comb regimes. We show that optimal computational performance occurs within the band of parametric resonance, while deeper frequency-comb regimes often exhibit degraded performance despite their increased spectral complexity. We trace this to its origin: the comb regimes gain dimensionality only at the cost of the state convergence reservoir computation depends on. See Secs.~II and~III of the supplementary material.

The coupled two-modal system consists of a directly driven mode at frequency $\omega_1$ (mode~1) and a parametrically excited lower-frequency mode at $\omega_2 \approx \omega_1/2$ (mode~2).
When the drive frequency $\omega_D \approx \omega_1$, intrinsic three-wave mixing mediates a parametric energy transfer from the driven mode to the subharmonic mode, provided the drive amplitude exceeds a threshold determined by the system damping and detuning. Above this threshold, the coupled system undergoes a Hopf bifurcation, leading to parametric resonance. Within a portion of this regime, periodic amplitude modulations arise, which manifest themselves in the spectral domain as a ladder of equally spaced, phase-coherent lines, i.e., a frequency comb \cite{ganesan_phononic_2017}.
The conditions for parametric resonance are analytically bounded by an Arnold tongue in drive amplitude--frequency space, whose boundaries are set by the resonance frequencies, quality factors, mode coupling strength, and drive detuning. Below this boundary, the system remains in static equilibrium; near the boundary it enters a parametric resonance regime; and well inside it exhibits sustained comb dynamics.
In addition, for combs to exist, the drive frequency should be sufficiently detuned from the resonance \cite{qi_existence_2020}. We explore this progression through distinct dynamical regimes: sub-threshold, parametric resonance, and combs for neuromorphic computation.

We derive the dynamics of the amplitudes of two coupled modes using slowly varying envelope approximation \cite{ganesan_phononic_2017,ganesan_excitation_2018,qi_existence_2020}; a self-contained derivation is given in Sec.~I of the supplementary material. The normalized complex amplitudes of the two modes, $\psi_1$ and $\psi_2$, evolve according to Eqs.~(\ref{eq:eq-1}) and (\ref{eq:eq-2}):
\begin{equation}
	\label{eq:eq-1}
	\frac{\partial \psi_1}{\partial \tau} = - i f - \left( 1 + i \Delta_1\right) \psi_1 + i \psi_2^2,
\end{equation}
\begin{equation}
	\label{eq:eq-2}
	\frac{\partial \psi_2}{\partial \tau} = - \left( \gamma_{21} + i \Delta_2\right)\psi_2 + 2i\psi_1 \psi_2^{*},
\end{equation}
Here, $\tau = \gamma_1 t$ is normalized time, $f$ is the dimensionless drive, $\Delta_1 \propto \omega_D-\omega_1$ and $\Delta_2  \propto \omega_D-2\omega_2$ are detunings, and $\gamma_{21} = \gamma_2/\gamma_1$ is the damping ratio. The detunings satisfy $\Delta_2 = \Delta_1/2 + \kappa$, where $\kappa$ is the detuning from the frequency matching condition $\omega_1=2\omega_2$. The nonlinear terms $i\psi_2^2$ and $2i\psi_1\psi_2^*$ mediate parametric energy exchange between modes. When $f \ge \tfrac{1}{2}|\gamma_{21}\Delta_1+\Delta_2|$, the system undergoes a Hopf bifurcation, producing parametric resonance. In addition, if $2\Delta_1\Delta_2 \le -(1+\Delta_1^2+2\gamma_{21})$, periodic amplitude modulations occur, which in turn generate a frequency comb in the spectral domain.

The parametric resonance/frequency comb reservoir operates as follows: A chaotic time series $s(n)$ is normalized to $[0,1]$. Each symbol $s(n)$ is then encoded into the normalized drive amplitude as:
\begin{equation}
    \label{eq:symbol-encoding}
    f(n) = \text{F}_{\text{avg}} + \Delta \text{F} \times \left( s(n) - 0.5 \right),
\end{equation}
$\text{F}_{\text{avg}}$ is the average drive setting the operating point within or near the comb regime and $\Delta \text{F}$ is the modulation depth. Unless stated otherwise we use the input-proportional depth $\Delta \text{F}=\text{F}_{\text{avg}}$, for which Eq.~(\ref{eq:symbol-encoding}) reduces to $f(n)=\text{F}_{\text{avg}}\left(s(n)+0.5\right)$, so the drive carries the input at every operating point; unlike a fixed $\Delta \text{F}$, this depth scales with $\text{F}_{\text{avg}}$ \mbox{\cite{shishavan_optical_2025,parto_ultrafast_2025}}.
Prior to the RC run, the reservoir is initialized by driving it at a constant amplitude $\text{F}_{\text{avg}}$ for 50 symbol periods, allowing the system to reach a quasi-steady state. For each input symbol, Eqs.~(\ref{eq:eq-1}) and (\ref{eq:eq-2}) are integrated using a fifth-order Runge--Kutta method over one symbol period $T_{\text{symbol}}$, corresponding to a data rate $\gamma_1/T_{\text{symbol}}$.

The high-dimensional reservoir state is read out via time-multiplexing with $N_v=512$ virtual nodes \cite{sun_novel_2021,guo_mems_2024}, sampled at equally spaced instants within each symbol period.
The feature vector for symbol $n$ combines mode~1 and mode~2 power trajectories with the first $N_{\mathrm{FFT}}=512$ Fourier magnitudes of $\psi_1$ and $\psi_2$. The power features discard phase, whereas the complex-amplitude spectra retain intra-symbol phase evolution and phase-sensitive harmonics. The spectral block therefore provides information complementary to the power trajectories rather than simply reproducing their frequency content. At parametric resonance, the combined features achieve test NMSE $4.51\times10^{-4}$, versus $3.31\times10^{-3}$ and $2.04\times10^{-3}$ for temporal and spectral features alone (Sec.~IV of the supplementary material).
The transforms of the complex modal amplitudes are evaluated over a single symbol period. Here, $k$ is the equally spaced sampling instants within each symbol period. The total feature dimensionality per symbol is, therefore, $2N_v + 2 N_{\text{FFT}}=2048$ (see Sec.~IV of the supplementary material). To accommodate the wide dynamic range of the modal amplitudes, all features $x$ are log-compressed as $\log_{10} \left(\left| x \right| + \epsilon \right)$, with $\epsilon=10^{-10}$, following which the feature matrix is standardized to zero mean and unit variance column-wise; features with zero variance across the training set are excluded.

The readout is trained via ridge regression regularization parameter, $\lambda=10^{-3}$ (see Sec.~VII of the supplementary material for the results when using the optimal values of $\lambda$ for each regime) to predict $s(n+1)$ \cite{lukosevicius_reservoir_2009,jaeger_harnessing_2004}. Prediction is open-loop throughout: the reservoir is driven by the measured series at every step, including during testing, and the predicted value is never fed back into the drive, so the reported errors are single-step accuracies that do not accumulate. The first 200 symbols are discarded to remove transients. Data are split in the ratio 80:20 for training/testing. The readout weights are obtained via the closed-form ridge regression solution shown in Eq.~(\ref{eq:ridge-eq}):
\begin{equation}
	\label{eq:ridge-eq}
	\boldsymbol{\beta} = \left( \mathbf{X}^\top \mathbf{X} + \lambda \mathbf{I} \right)^{-1} \mathbf{X}^\top \mathbf{y},
\end{equation}
where $\mathbf{X} \in \mathbb{R}^{N \times D}$ is the feature matrix with $D=2048$ features per symbol and $N$ the number of training samples, $\mathbf{y} \in \mathbb{R}^{N}$ is the corresponding target vector containing the next-step values $s(n+1)$.
\begin{figure*}
	\centering
	\includegraphics[width=1.0\textwidth]{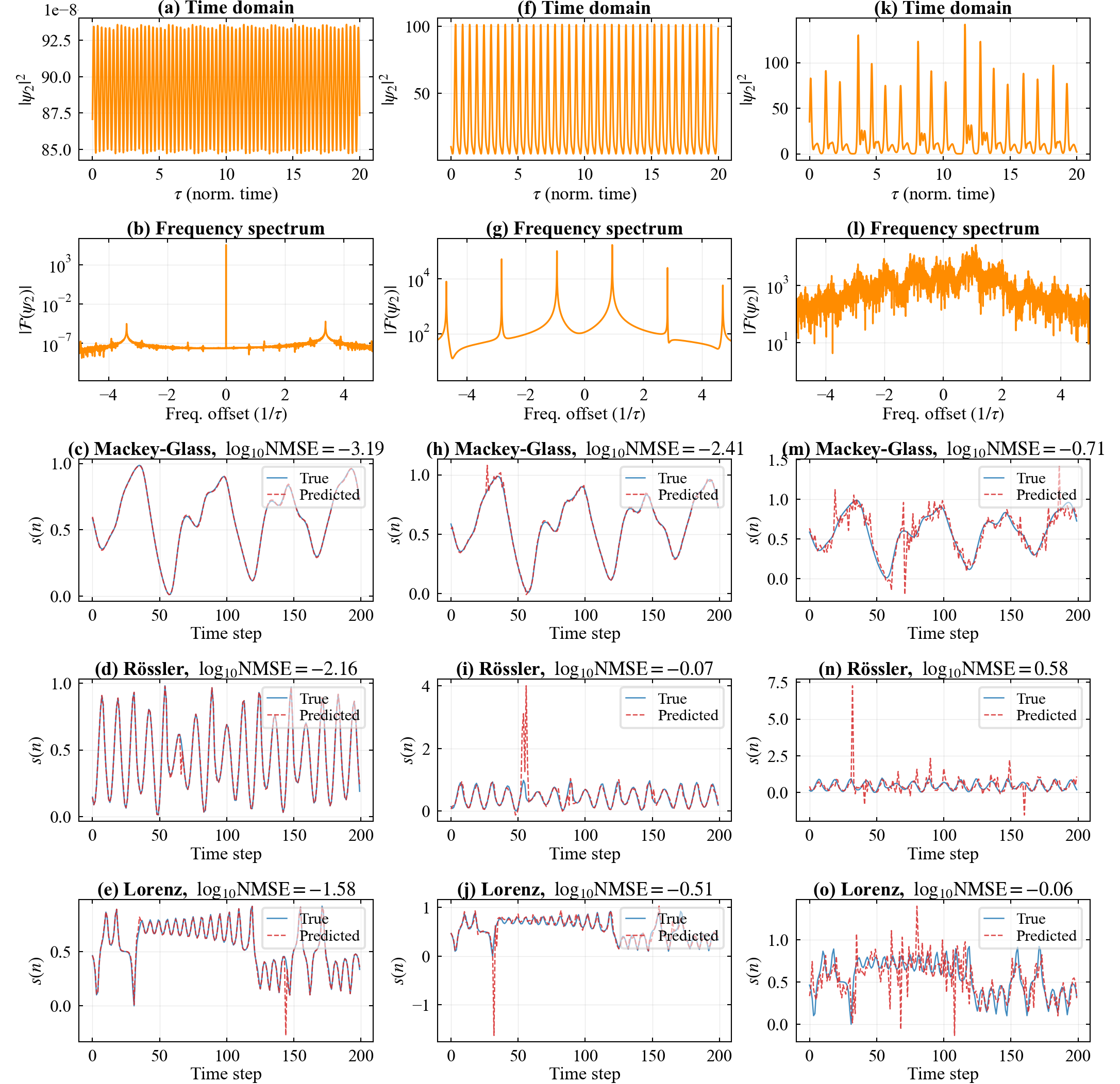}
	\caption{\label{fig:all-preds} Time domain, frequency domain and prediction performance of Mackey--Glass, Rössler, and Lorenz chaotic time series of (a-e) Parametric resonance ($\text{F}_\text{avg}=40.0,\ \Delta_1=-4.5$), (f-j) Coherent comb region ($\text{F}_\text{avg}=40.0,\ \Delta_1=+4.5$) and (k-o) Chaotic comb region ($\text{F}_\text{avg}=20.0,\ \Delta_1=+2.0$). The prediction curves show open-loop one-step-ahead prediction: the measured input is supplied to the reservoir at every step during both training and testing, and the predicted output is never fed back into the drive. Note: $\Delta \text{F}=\text{F}_{\text{avg}}$,~$\gamma_{21}=1.0$, $\kappa=-9$ and input data rate = 2424 samples/second.}
\end{figure*}
The performance of the reservoir is quantified using the normalized mean squared error (NMSE), defined as in Eq.~(\ref{eq:nmse-calc}):
\begin{equation}
	\label{eq:nmse-calc}
	\text{NMSE} = \frac{\sum_{n} \left( s(n) - \hat{s}(n) \right)^2}{\sum_{n} \left( s(n) - \bar{s} \right)^2},
\end{equation}
where $\bar{s}$ denotes the mean of the ground-truth signal. Lower NMSE indicates improved performance.

We vary $\text{F}_{\text{avg}}$, $\Delta \text{F}$, $\Delta_1$ and $\gamma_{21}$ across the Hopf boundary, identifying: (i) sub-threshold (no parametric excitation), (ii) parametric resonance (near-threshold), and (iii) the comb regime. Eqs.~(\ref{eq:eq-1}) and (\ref{eq:eq-2}) are integrated with adaptive time stepping, with relative and absolute tolerances of $10^{-8}$ and $10^{-10}$, requiring between 508 and 5295 right-hand-side evaluations per symbol depending on the regime. Convergence was verified against temporal resolution and feature sampling. Tolerance convergence, and the absence of stiffness in all regimes examined, are established in Sec.~VIII of the supplementary material. Memory is controlled via $T_{\text{symbol}}$ relative to $\gamma_1^{-1}$. For $T_{\text{symbol}} \gg \gamma_1^{-1}$, inputs decouple; for $T_{\text{symbol}} \ll \gamma_1^{-1}$, the system cannot respond. Optimal performance occurs when residual energy overlaps successive inputs.

\begin{figure*}[t]
	\centering
	\includegraphics[width=1.0\textwidth]{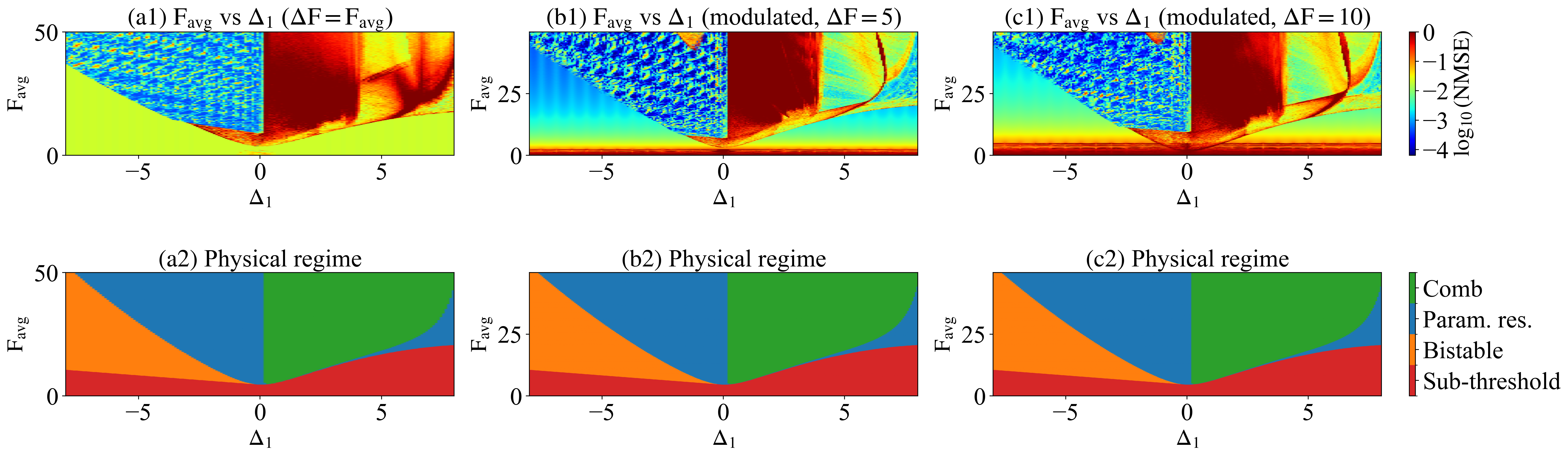}
	\caption{\label{fig:f_avg_vs_delta1}
		(a-c) Normalized mean squared error (NMSE, log scale) as a function of average drive amplitude $\text{F}_{\mathrm{avg}}$ and detuning $\Delta_1$, along with the corresponding dynamical regime maps showing sub-threshold (ST), parametric resonance (PR), and frequency comb (FC) regions for modulation depths $\Delta \text{F} = \text{F}_{\text{avg}}$, $5$, and $10$. The lowest NMSE values are concentrated within the parametric resonance regime, while heterogeneous performance is observed across the comb regime. The comb region is resolved into coherent and chaotic sub-regions at constant drive in Sec.~V of the supplementary material. Note: $\gamma_{21}=1.0$, $\kappa=-9$ and input data rate = 2424 samples/second.
	}
\end{figure*}

We evaluated our parametric resonance/frequency comb reservoir on the Mackey--Glass, Rössler, and Lorenz systems using open-loop one-step-ahead linear readout prediction over 1000 time steps. The measured input is supplied at each step, with no feedback of the predicted output into the drive.

\begin{figure*}[t]
	\centering
	\includegraphics[width=1.0\textwidth]{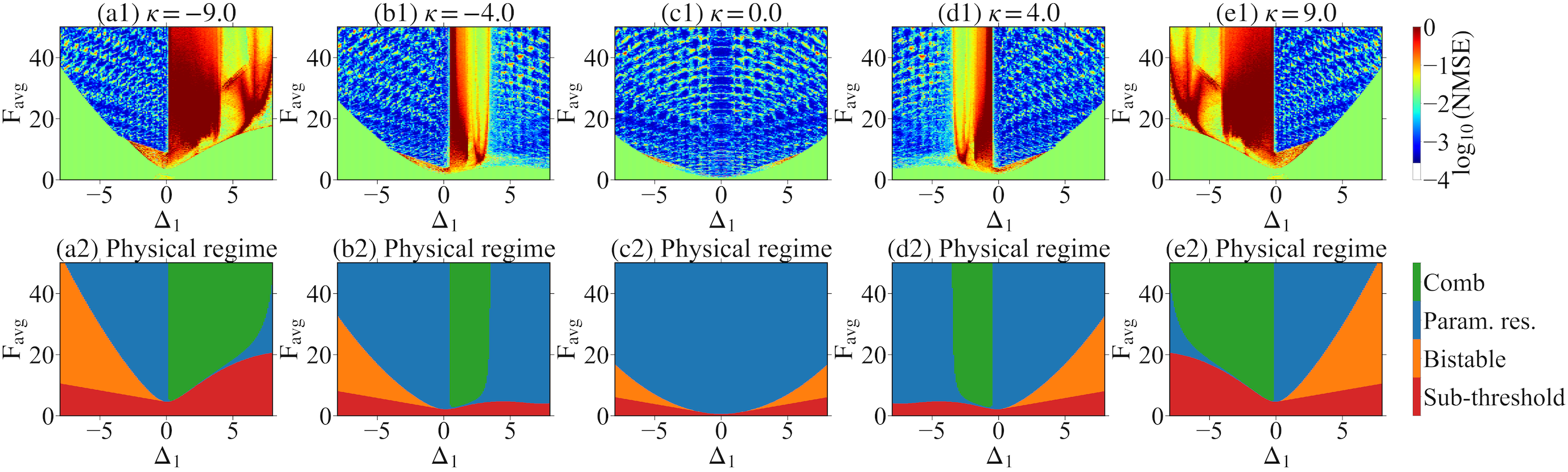}
	\caption{\label{fig:kappa-sweep}
		(a-e) Dependence of prediction performance on the detuning offset $\kappa$. NMSE (log scale) is shown over the $(\text{F}_{\mathrm{avg}}, \Delta_1)$ parameter space for different values of $\kappa$, along with the corresponding dynamical regime maps. Variation in $\kappa$ reshapes the parametric resonance and comb regions, with the lowest NMSE consistently aligned with the parametric resonance regime. Note: $\Delta \text{F}=\text{F}_{\text{avg}}$, $\gamma_{21}=1.0$ and input data rate = 2424 samples/second.}
\end{figure*}
In the parametric resonance regime (Figs.~\ref{fig:all-preds}(a-e)), the time domain response of $\left| \psi_2 \right|^2$ (Fig.~\ref{fig:all-preds}(a)) exhibits regular, periodic oscillations with a spectrally sparse response. These structured dynamics translate into strong computational performance, with low prediction errors across the Mackey--Glass (Fig.~\ref{fig:all-preds}(c); $\log_{10}(\text{NMSE}) = -3.19$), Rössler (Fig.~\ref{fig:all-preds}(d); $\log_{10}(\text{NMSE}) = -2.16$), and Lorenz (Fig.~\ref{fig:all-preds}(e); $\log_{10}(\text{NMSE}) = -1.58$) chaotic time series.
Note: Although the NMSE was calculated for over 1000 time steps, Fig.~\mbox{\ref{fig:all-preds}} shows the time series for only up to 200 time steps for legibility.
\begin{figure*}[t]
	\centering
	\includegraphics[width=1.0\textwidth]{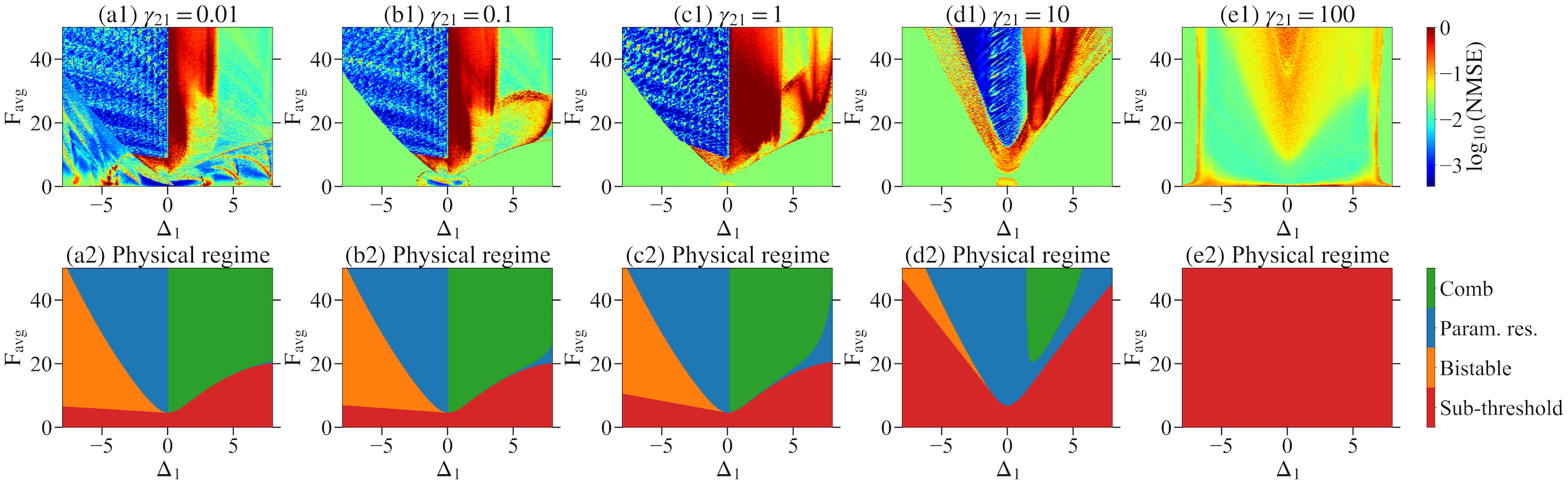}
	\caption{\label{fig:gamma21-sweep}
		(a-e) Dependence of prediction performance on the damping ratio $\gamma_{21}$. NMSE (log scale) is shown over the $(\text{F}_{\mathrm{avg}}, \Delta_1)$ parameter space for different values of $\gamma_{21}$, along with the corresponding dynamical regime maps (a2-e2). Increasing $\gamma_{21}$ suppresses the parametric instability and reduces the extent of the low-NMSE region. Note: $\Delta \text{F}=\text{F}_{\text{avg}}$, $\kappa=-9$ and input data rate = 2424 samples/second.
	}
\end{figure*}
In the coherent comb region (Figs.~\ref{fig:all-preds}(f-j)), periodic amplitude modulation leads to the formation of equally spaced spectral lines, reflecting phase-coherent multi-frequency dynamics. While this regime increases the dimensionality of the reservoir through harmonic generation, the prediction performance degrades relative to parametric resonance. The errors increase to $\log_{10}(\text{NMSE}) =-2.41,\ -0.07,\ -0.51$ for the Mackey--Glass, Rössler and Lorenz chaotic time series, respectively (Figs.~\ref{fig:all-preds}(h-j)).
\begin{figure*}[t]
	\centering
	\includegraphics[width=1.0\textwidth]{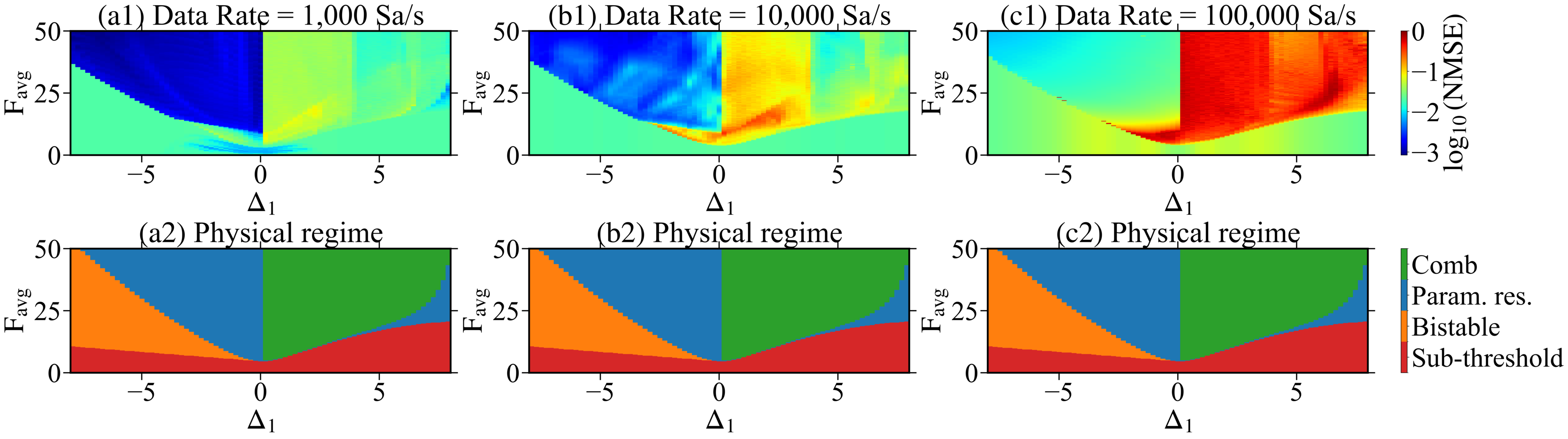}
	\caption{\label{fig:data-rate}
		(a-c) Dependence of prediction performance on input data rate. NMSE (log scale) is shown over the $(\text{F}_{\mathrm{avg}}, \Delta_1)$ parameter space for different data rates. Optimal performance is observed at intermediate data rates, where the input timescale matches the intrinsic dynamical response of the system. Note: $\Delta \text{F}=\text{F}_{\text{avg}}$, $\gamma_{21}=1.0$ and $\kappa=-9$.
	}
\end{figure*}
In the chaotic comb regime (Figs.~\ref{fig:all-preds}(k-o)), the time-domain signal becomes irregular and strongly fluctuating, and the frequency spectrum broadens into a continuous distribution, indicating the loss of phase coherence. In this regime, the prediction performance deteriorates significantly, with $\log_{10}(\text{NMSE}) = -0.71,\ 0.58,\ -0.06$ for the Mackey--Glass, Rössler and Lorenz chaotic time series, respectively (Figs.~\ref{fig:all-preds}(m-o)).
The reconstructed trajectories show large deviations and intermittent instabilities, reflecting the breakdown of the structured nonlinear transformation required for effective reservoir computation. This is explained in Secs.~II and~III of the supplementary material.

We evaluated information processing capacity (IPC) \cite{dambre_information_2012}, effective dimensionality, and the echo-state property across regimes (Secs.~II and III of the supplementary material). IPC falls from $160.3$ at parametric resonance to $18.0$ in the coherent comb and $5.5$ in the chaotic comb. These are lower bounds due to the degree-11 polynomial and finite-delay truncation and the parametric-resonance capacity remains increasing at the highest degree. Its per-degree capacity reaches $34.1$, versus $0.26$ in the chaotic comb, indicating that computational differences reflect the depth of nonlinear transformations, not merely state dimensionality.

The echo-state property and effective dimensionality $D_{\mathrm{eff}}$ capture distinct aspects of reservoir dynamics. $D_{\mathrm{eff}}$ measures variance across principal directions, while the echo-state property requires convergence under identical inputs. $D_{\mathrm{eff}}$ rises from $1.91$ in the sub-threshold regime to $2.77$, $7.15$, and $7.40$ at parametric resonance, coherent comb, and chaotic comb, respectively, while the convergence rate falls from $4.03$ per symbol to approximately zero in both comb regimes. Thus, greater dimensionality does not necessarily yield computationally useful representations.

The effective dimensionality $D_{\mathrm{eff}}$ is derived from covariance eigenvalues and is therefore not equivalent to the number of spectral comb teeth. Comb teeth represent distinct frequency components, whereas $D_{\mathrm{eff}}$ measures variance across principal directions. Correlated or weakly excited spectral components can therefore increase spectral richness without proportionally increasing effective dimensionality or readout performance.

The comb regimes are separated by the largest Lyapunov exponent at constant drive \cite{benettin_lyapunov_1980}.
The threshold $\lambda_{\max}>10^{-2}$ exceeds the finite-time estimator scatter, with median $2.3\times10^{-4}$ and 99th percentile $7.3\times10^{-3}$. The classification remains robust, with chaotic-comb fractions of 45.9\%, 45.8\%, 45.7\%, and 45.4\% for thresholds $3\times10^{-3}$, $10^{-2}$, $3\times10^{-2}$, and $10^{-1}$, respectively (see the supplementary material).
The measured exponents are $-0.007$, $+3\times10^{-5}$, and $+0.563$ per $\tau$ at the parametric resonance, coherent comb, and chaotic comb points, respectively. The sign change defines the boundary mapped over the $\left(\text{F}_\text{avg},\Delta_1\right)$ plane in Sec.~V of the supplementary material.

Fig.~\ref{fig:f_avg_vs_delta1}(a1) maps the prediction error over the $\left( \text{F}_\text{avg},\ \Delta_1 \right)$ parameter space for Mackey--Glass, while Fig.~\ref{fig:f_avg_vs_delta1}(a2) shows the corresponding dynamical regime classification. Extended parameter sweeps for the Lorenz and Rössler systems are provided in the supplementary material. The lowest NMSE values are concentrated within the parametric resonance regime, establishing a direct correspondence between computational performance and the underlying bifurcation structure.

A bistable region is observed between the sub-threshold and parametric resonance regimes, as shown in Fig.~\ref{fig:f_avg_vs_delta1}(b). In this region, both the trivial steady state ($\psi_2=0$) and the parametric state ($\psi_2\neq0$) are stable. The boundaries of this region are determined by the onset of existence and stability of the parametric branch, with the lower boundary given by $\text{f}_{\text{Arnold}} = \frac{1}{2} \left| \gamma_{21}\Delta_1+\Delta_2\right|$ and the upper bound by $\text{f}_\text{upper} = \frac{1}{2} \sqrt{\left(1+\Delta_1^2\right)\left(\gamma_{21}^2+\Delta_2^2\right)}$ \cite{qi_existence_2020}. The computational performance observed in Fig.~\ref{fig:f_avg_vs_delta1}(a) within this region corresponds to the trivial branch, as the reservoir is initialized in this state. The final state depends on the initial conditions. Consequently, the low NMSE in the bistable region reflects initialization-dependent operation rather than an intrinsically robust regime. Over 32 initial conditions, the constant-amplitude startup never selects the parametric branch in any of the trials. However, a drive ramped from above $\text{f}_\text{upper}$ selects the parametric branch in all trials (Sec.~VI of the supplementary material).

The effect of modulation depth $\Delta\text{F}$ on reservoir performance was studied.
Replacing the input-proportional depth by a fixed $\Delta\text{F}$ removes its growth along the $\text{F}_{\text{avg}}$ axis, and the low-NMSE region corresponding to parametric resonance becomes more uniform (Figs.~\ref{fig:f_avg_vs_delta1}(b) and \ref{fig:f_avg_vs_delta1}(c)). While a degradation is observed at very low $\text{F}_{\mathrm{avg}}$, where the system is intermittently driven into the sub-threshold regime, the overall performance improves at higher drive strengths. Furthermore, both the coherent and chaotic comb regimes exhibit improved prediction accuracy with increasing fixed $\Delta\text{F}$, indicating that modulation enhances the robustness of the nonlinear transformation, which in turn strengthens the prediction performance of the system.

The effect of the detuning $\kappa$ on the reservoir performance is shown in Fig.~\ref{fig:kappa-sweep}.
Varying $\kappa$ reshapes the parametric resonance and comb regions in the $\left(\text{F}_\text{avg},\ \Delta_1\right)$ space, altering both their extent and symmetry.
The region of lowest NMSE consistently follows the parametric resonance regime as it shifts with $\kappa$, indicating that optimal computation remains tied to this bifurcation structure. For large values of $\left| \kappa \right|$, the resonance region becomes distorted and reduced, leading to a shrinkage of the low-error region. This shows that $\kappa$ directly controls the accessible nonlinear operating regime and, consequently, the computational performance.

Fig.~\ref{fig:gamma21-sweep} shows the impact of the damping ratio $\gamma_{21}$ on the reservoir dynamics. Increasing $\gamma_{21}$ suppresses the parametric instability and progressively reduces the extent of both the parametric resonance and comb regimes. As a result, the low-NMSE region contracts and eventually disappears at large values of $\gamma_{21}$, where the system is confined to sub-threshold dynamics. In contrast, for small $\gamma_{21}$, the parametric region expands, but the performance becomes less uniform due to the increased sensitivity to parameter variations. The best performance remains localized in the parametric resonance regime, confirming that a balance between damping and nonlinear coupling is required for optimal prediction.

The dependence of performance on input data rate is shown in Fig.~\ref{fig:data-rate}. At low data rates, the symbol duration is long compared to the system relaxation time, allowing the dynamics to settle between inputs and resulting in poor performance due to loss of memory. At intermediate data rates, a well-defined low-NMSE region emerges within the parametric resonance regime, where residual dynamics overlap successive inputs, enabling both memory and nonlinear transformation. At high data rates, the system cannot fully respond within a symbol period, leading to degraded performance and a reduction of the low-error region.
Although driven at a common rate, the three benchmarks are subsampled beforehand to comparable dynamical footing, at $17.0$, $7.5$ and $12.0$ samples per characteristic time for Mackey--Glass, Lorenz and Rössler.
Additional data-rate sweeps and sparse LASSO-based readout results are provided in the supplementary material.

These results demonstrate that the computational capability of parametrically driven reservoirs is governed by the underlying bifurcation structure.
These findings establish parametric resonance as a robust operating regime for neuromorphic computation and provide design guidelines for tuning physical oscillator systems toward optimal performance.
Future work will explore higher-dimensional systems while addressing noise, fabrication variability, bandwidth, and branch-selection. In the bistable regime, constant-amplitude startup can remain on the trivial branch, whereas ramping above the upper stability boundary selects the parametric branch, making robust initialization or active control important for MEMS and phononic comb implementations toward ultrafast, energy-efficient neuromorphic hardware.

	\section*{Supplementary Material}
	See the supplementary material for the derivation of the two-oscillator equations; IPC decomposition; echo-state and effective-dimensionality analyses; feature-vector, ablation, and principal-angle analyses; the Lyapunov criterion and regime map; bistable branch selection; readout regularization; numerical convergence and stiffness tests; and additional Lorenz and Rössler sweeps, input-rate dependence, and LASSO readouts.

	\begin{acknowledgments}
		The authors gratefully acknowledge the computing time provided on the high performance computing facility, Sharanga, at the Birla Institute of Technology and Science (BITS) Pilani, Hyderabad Campus.
	\end{acknowledgments}

	\bibliography{bibliography}

\onecolumngrid
\clearpage
\makeatletter
\def\title@column#1{%
 \minipagefootnote@init
 \begingroup
  \let\@footnotetext\frontmatter@footnotetext
  \ltx@no@footnote
  #1%
 \endgroup
 \minipagefootnote@foot
}%
\let\set@footnotewidth\set@footnotewidth@one
\onecolumn@grid@setup
\expandafter\let\csname and\expandafter\endcsname\csname supp@sv@and\endcsname
\expandafter\let\csname affiliation\expandafter\endcsname\csname supp@sv@affiliation\endcsname
\expandafter\let\csname author\expandafter\endcsname\csname supp@sv@author\endcsname
\expandafter\let\csname @AF@join\expandafter\endcsname\csname supp@sv@@AF@join\endcsname
\expandafter\let\csname email\expandafter\endcsname\csname supp@sv@email\endcsname
\expandafter\let\csname @address\expandafter\endcsname\csname supp@sv@@address\endcsname
\expandafter\let\csname maketitle\expandafter\endcsname\csname supp@sv@maketitle\endcsname
\expandafter\let\csname thanks\expandafter\endcsname\csname supp@sv@thanks\endcsname
\frontmatter@init
\makeatother
\setcounter{section}{0}
\setcounter{equation}{0}
\setcounter{figure}{0}
\setcounter{table}{0}

\title[Supplemental Material]{Supplemental Material for: Neuromorphic Computing Based on Parametrically-Driven Oscillators and Frequency Combs}

\author{Mahadev Sunil Kumar}
	\email{mahadevsunilkumar03@gmail.com}
	\homepage{https://www.mahadevsunilkumar.com}
	\affiliation{Amrita School of Computing, Amrita Vishwa Vidyapeetham, Kollam, India 690525}
	\author{Adarsh Ganesan}%
	\email{adarsh@dubai.bits-pilani.ac.in}
	\homepage{https://www.bits-pilani.ac.in/dr-adarsh-ganesan}
	\affiliation{Department of Electrical and Electronics Engineering, Birla Institute of Technology and Science, Pilani -- Dubai Campus, Dubai, United Arab Emirates 345055}
\date{\today}

\maketitle
\setcounter{page}{1}

This supplementary material provides the derivation of the coupled-mode model, explains the performance differences between the dynamical regimes, and gives the numerical checks supporting the main results. We first derive Eqs.~\ref{eq:35}--\ref{eq:36} from the two-mode equations of motion. We then examine why the frequency comb regime underperforms despite its richer dynamics. An information processing capacity decomposition shows which computational resources are provided by each regime, while measurements of effective dimensionality and the echo-state property show that the comb regimes gain dimensionality at the expense of state convergence. We then describe the measurements used in the analysis, including the independence of the time domain and spectral feature blocks, the criterion used to distinguish coherent and chaotic combs, the branch selected in the bistable region, the cross-validated choice of readout regularization, and the convergence and stiffness tests for the integrator. Finally, we test the robustness of the results using additional chaotic benchmarks, input data rates, and readout regressions.

\section{Derivation of dynamics of the amplitudes of two coupled
modes}
\label{sec:derivation}
The equations of motion corresponding to two coupled modes are given by:
\begin{equation}
	\ddot{x}_1 + 2\gamma_1 \dot{x}_1 + \omega_1^2x_1 + \alpha_{22}x_2^2 = F \cos \left( \omega_D t \right)
	\label{eq:1}
\end{equation}
\begin{equation}
	\ddot{x}_2 + 2 \gamma_2 \dot{x}_2 + \omega_2^2 x_2 + \alpha_{12}x_1x_2=0
	\label{eq:2}
\end{equation}

The modes are written as
\begin{equation}
	x_1 = \frac{1}{2} \left( u_1 e^{i \omega_D t} + u_1^* e^{-i \omega_D t} \right)
	\label{eq:3}
\end{equation}
\begin{equation}
	x_2 = \frac{1}{2} \left( u_2 e^{i
	\frac{\omega_D}{2}t}
	+ u_2^*e^{
	-i \frac{\omega_D}{2}t}
	 \right)
	 \label{eq:4}
\end{equation}

The derivatives $\dot{x_1}$, $\dot{x_2}$, $\ddot{x_1}$, and $\ddot{x_2}$ and the coupling terms $x_1x_2$, and $x_2^2$ are obtained as:
\begin{equation}
	\dot{x}_1 = \frac{1}{2} \left(
	i \omega_D u_1 e^{i \omega_D t} + \dot{u}_1 e^{i \omega_D t} - i \omega_D u_1 e^{-i \omega_D t} + \dot{u}_1^*e^{-i \omega_D t}
	\right)
	\label{eq:5}
\end{equation}
\begin{equation}
	\dot{x}_2 = \frac{1}{2} \left(
	i \frac{\omega_D}{2}u_2 e^{i \frac{\omega_D}{2}t} + \dot{u}_2 e^{i \frac{\omega_D}{2}t} - i \frac{\omega_D}{2} u_2 e^{-i \frac{\omega_D}{2}t} + \dot{u}_2^* e^{-i \frac{\omega_D}{2}t}
	\right)
	\label{eq:6}
\end{equation}
\begin{equation}
	\ddot{x}_1 = \frac{1}{2} \left(
	\ddot{u}_1 e^{i \omega_D t} + 2i \omega_D \dot{u}_1 e^{i \omega_D t} - \omega_D^2 u_1 e^{i \omega_D t} + \ddot{u}_1^* e^{-i \omega_D t} - 2 i \omega_D \dot{u}_1^* e^{-i \omega_D t} - \omega_D^2 u_1^* e^{-i \omega_D t}
	\right)
	\label{eq:7}
\end{equation}
\begin{equation}
	\ddot{x}_2 = \frac{1}{2} \left(
	\ddot{u}_2 e^{i \frac{\omega_D}{2}t} + i \omega_D \dot{u}_2 e^{i \frac{\omega_D}{2}t} - \frac{\omega_D^2}{4} u_2 e^{i \frac{\omega_D}{2}t} + \ddot{u}_2^* e^{-i \frac{\omega_D}{2}t} - i \omega_D \dot{u}_2^* e^{-i \frac{\omega_D}{2}t} - \frac{\omega_D^2}{4} u_2^* e^{-i \frac{\omega_D}{2}t}
	\right)
	\label{eq:8}
\end{equation}
\begin{equation}
	x_1 x_2 = \frac{1}{4} \left(
	u_1^* u_2^* e^{-3i \frac{\omega_D}{2}t} + u_1^* u_2 e^{-i \frac{\omega_D}{2}t} + u_1 u_2^* e^{i \frac{\omega_D}{2}t} + u_1 u_2 e^{3i \frac{\omega_D}{2}t}
	\right)
	\label{eq:9}
\end{equation}
\begin{equation}
	x_2^2 = \frac{1}{4}\left(u_2^2 e^{i\omega_D t} + {u_2^*}^2 e^{-i\omega_D t} + 2|u_2|^2\right)
	\label{eq:10}
\end{equation}

Now, by taking $\cos(\omega_D t) = \frac{1}{2}\left(e^{i\omega_D t} + e^{-i\omega_D t}\right)$, Eq.~\ref{eq:1} and \ref{eq:2} becomes:

\begin{equation}
	\begin{aligned}
		\frac{1}{2}\Big(&
		\ddot{u}_1 e^{i\omega_D t}
		+ 2i\omega_D \dot{u}_1 e^{i\omega_D t}
		- \omega_D^2 u_1 e^{i\omega_D t}
		+ \ddot{u}_1^* e^{-i\omega_D t}
		- 2i\omega_D \dot{u}_1^* e^{-i\omega_D t}
		- \omega_D^2 u_1^* e^{-i\omega_D t}\Big) \\
		&+ \gamma_1\Big(
		i\omega_D u_1 e^{i\omega_D t}
		+ \dot{u}_1 e^{i\omega_D t}
		- i\omega_D u_1 e^{-i\omega_D t}
		+ \dot{u}_1^* e^{-i\omega_D t}\Big) \\
		&+ \frac{\omega_1^2}{2}\Big(
		u_1 e^{i\omega_D t}
		+ u_1^* e^{-i\omega_D t}\Big)
		+ \frac{\alpha_{22}}{4}\Big(
		u_2^2 e^{i\omega_D t}
		+ {u_2^*}^2 e^{-i\omega_D t}
		+ 2|u_2|^2\Big)
		= \frac{F}{2}\Big(e^{i\omega_D t}+e^{-i\omega_D t}\Big)
		\label{eq:11}
	\end{aligned}
\end{equation}

\begin{equation}
	\begin{aligned}
		\frac{1}{2}\Big(&
		\ddot{u}_2 e^{i\frac{\omega_D}{2}t}
		+ i\omega_D\dot{u}_2 e^{i\frac{\omega_D}{2}t}
		- \frac{\omega_D^2}{4}u_2 e^{i\frac{\omega_D}{2}t}
		+ \ddot{u}_2^* e^{-i\frac{\omega_D}{2}t}
		- i\omega_D\dot{u}_2^* e^{-i\frac{\omega_D}{2}t}
		- \frac{\omega_D^2}{4}u_2^* e^{-i\frac{\omega_D}{2}t}\Big) \\
		&+ \gamma_2\Big(
		i\frac{\omega_D}{2}u_2 e^{i\frac{\omega_D}{2}t}
		+ \dot{u}_2 e^{i\frac{\omega_D}{2}t}
		- i\frac{\omega_D}{2}u_2 e^{-i\frac{\omega_D}{2}t}
		+ \dot{u}_2^* e^{-i\frac{\omega_D}{2}t}\Big) \\
		&+ \frac{\omega_2^2}{2}\Big(
		u_2 e^{i\frac{\omega_D}{2}t}
		+ u_2^* e^{-i\frac{\omega_D}{2}t}\Big)
		+ \frac{\alpha_{12}}{4}\Big(
		u_1^*u_2^*e^{-3i\frac{\omega_D}{2}t}
		+ u_1^*u_2e^{-i\frac{\omega_D}{2}t}
		+ u_1u_2^*e^{i\frac{\omega_D}{2}t}
		+ u_1u_2e^{3i\frac{\omega_D}{2}t}\Big)
		= 0
		\label{eq:12}
	\end{aligned}
\end{equation}

We now collect the coefficients of $e^{i\omega_D t}$ in Eq.~\ref{eq:11} and coefficients of $e^{i \frac{\omega_D}{2} t}$ in Eq.~\ref{eq:12} to obtain the following relationships.

\begin{equation}
	\ddot{u}_1 + \left[2\gamma_1 + 2i\omega_D\right]\dot{u}_1 + \left[(\omega_1^2 - \omega_D^2) + 2i\gamma_1 \omega_D\right]u_1 + \alpha_{22}\frac{u_2^2}{2} = F
	\label{eq:13}
\end{equation}
\begin{equation}
	\ddot{u}_2 + \left[2\gamma_2 + i\omega_D\right]\dot{u}_2 + \left[\left(\omega_2^2 - \frac{\omega_D^2}{4}\right) + i\gamma_2 \omega_D\right]u_2 + \frac{\alpha_{12}}{2} u_1 u_2^* = 0
	\label{eq:14}
\end{equation}

We now multiply Eq.~\ref{eq:13} by $\frac{\alpha_{12}}{2\omega_D^4}$ and Eq.~\ref{eq:14} by $\frac{\sqrt{\alpha_{12}\alpha_{22}}}{2\omega_D^4}$.
\begin{equation}
	\frac{\partial^2 \left(\frac{\alpha_{12}}{2\omega_D^2} u_1\right)}{\partial(\omega_D t)^2} + \left[2\left(\frac{\gamma_1}{\omega_D}\right) + 2i\right]\dot{u}_1 + \left[\left(\frac{\omega_1}{\omega_D}\right)^2 - 1 + 2i\left(\frac{\gamma_1}{\omega_D}\right)\right]\frac{\partial \left(\frac{\alpha_{12}}{2\omega_D^2} u_1\right)}{\partial(\omega_D t)} + \left(\frac{\sqrt{\alpha_{12}\alpha_{22}}}{2\omega_D^2} u_2\right)^2 = \frac{\alpha_{12}}{2\omega_D^4} F
	\label{eq:15}
\end{equation}
\begin{equation}
	\frac{\partial^2 \left(\frac{\sqrt{\alpha_{12}\alpha_{22}}}{2\omega_D^2} u_2\right)}{\partial(\omega_D t)^2} + \left[2\left(\frac{\gamma_2}{\omega_D}\right) + i\right]\dot{u}_2 + \left[\left(\frac{\omega_2}{\omega_D}\right)^2 - \frac{1}{4} + i\left(\frac{\gamma_2}{\omega_D}\right)\right]\frac{\partial \left(\frac{\sqrt{\alpha_{12}\alpha_{22}}}{2\omega_D^2} u_2\right)}{\partial(\omega_D t)} + \left(\frac{\alpha_{12}}{2\omega_D^2} u_1\right)\left(\frac{\sqrt{\alpha_{12}\alpha_{22}}}{2\omega_D^2} u_2^*\right) = 0
	\label{eq:16}
\end{equation}
Now, we apply the following transformation and hence obtain Eq.~\ref{eq:18} and Eq.~\ref{eq:19}.

\begin{equation}
	\begin{aligned}
		T &= \omega_D t, \quad
		\bar{F} = \frac{\alpha_{12}}{2\omega_D^4} F \\
		\bar{\psi}_1 &= \frac{\alpha_{12}}{2\omega_D^2} u_1, \quad
		\bar{\psi}_2 = \frac{\sqrt{\alpha_{12}\alpha_{22}}}{2\omega_D^2} u_2 \\
		\bar{\omega}_1 &= \frac{\omega_1}{\omega_D}, \quad
		\bar{\omega}_2 = \frac{\omega_2}{\omega_D} \\
		\bar{\gamma}_1 &= \frac{\gamma_1}{\omega_D}, \quad
		\bar{\gamma}_2 = \frac{\gamma_2}{\omega_D}
	\end{aligned}
	\label{eq:17}
\end{equation}
\begin{equation}
	\frac{\partial^2 \bar{\psi}_1}{\partial T^2} + \left[2\bar{\gamma}_1 + 2i\right]\frac{\partial \bar{\psi}_1}{\partial T} + \left[\bar{\omega}_1^2 - 1 + 2i\bar{\gamma}_1\right]\bar{\psi}_1 + \bar{\psi}_2^2 = \bar{F}
	\label{eq:18}
\end{equation}
\begin{equation}
	\frac{\partial^2 \bar{\psi}_2}{\partial T^2} + \left[2\bar{\gamma}_2 + i\right]\frac{\partial \bar{\psi}_2}{\partial T} + \left[\bar{\omega}_2^2 - \frac{1}{4} + i\bar{\gamma}_2\right]\bar{\psi}_2 + \bar{\psi}_1 \bar{\psi}_2^* = 0
	\label{eq:19}
\end{equation}
For slowly varying fields, we assume the following conditions and hence we obtain Eq.~\ref{eq:23} and Eq.~\ref{eq:24}.
\begin{equation}
	\frac{\partial^2}{\partial T^2} \to 0
	\label{eq:20}
\end{equation}
\begin{equation}
	\bar{\gamma}_1 \frac{\partial}{\partial T} \to 0, \quad \bar{\gamma}_2 \frac{\partial}{\partial T} \to 0
	\label{eq:21}
\end{equation}
\begin{equation}
	\bar{\omega}_1 \to 1, \quad \bar{\omega}_2 \to \frac{1}{2}
	\label{eq:22}
\end{equation}
\begin{equation}
	2i\frac{\partial \bar{\psi}_1}{\partial T} + 2i\bar{\gamma}_1 \bar{\psi}_1 + \bar{\psi}_2^2 = \bar{F}
	\label{eq:23}
\end{equation}
\begin{equation}
	i\frac{\partial \bar{\psi}_2}{\partial T} + i\bar{\gamma}_2 \bar{\psi}_2 + \bar{\psi}_1 \bar{\psi}_2^* = 0
	\label{eq:24}
\end{equation}

We now multiply Eq.~\ref{eq:23} by $\frac{1}{4 \bar{\gamma}_1^2}$ and Eq.~\ref{eq:24} by $\frac{1}{2\bar{\gamma}_1^2}$.
\begin{equation}
	\frac{\partial \left(\frac{\bar{\psi}_1}{2\bar{\gamma}_1}\right)}{\partial(\bar{\gamma}_1 T)} = -\frac{\bar{\psi}_1}{2\bar{\gamma}_1} + i\left(\frac{\bar{\psi}_2}{2\bar{\gamma}_1}\right)^2 - i\frac{\bar{F}}{4\bar{\gamma}_1^2}
	\label{eq:25}
\end{equation}
\begin{equation}
	\frac{\partial \left(\frac{\bar{\psi}_2}{2\bar{\gamma}_1}\right)}{\partial(\bar{\gamma}_1 T)} = -\frac{\bar{\gamma}_2}{\bar{\gamma}_1}\left(\frac{\bar{\psi}_2}{2\bar{\gamma}_1}\right) + 2i\left(\frac{\bar{\psi}_1}{2\bar{\gamma}_1}\right)\left(\frac{\bar{\psi}_2^*}{2\bar{\gamma}_1}\right)
	\label{eq:26}
\end{equation}
Now, we make the following transformations.
\begin{equation}
	\tau = \bar{\gamma}_1 T; \quad f = \frac{\bar{F}}{4\bar{\gamma}_1^2}; \quad \gamma_{21} = \frac{\bar{\gamma}_2}{\bar{\gamma}_1}; \quad \kappa = \frac{\bar{\omega}_1 - 2\bar{\omega}_2}{2\bar{\gamma}_1}
	\label{eq:27}
\end{equation}
\begin{equation}
	\Delta_1 = \frac{1 - \bar{\omega}_1}{\bar{\gamma}_1}; \quad \Delta_2 = \frac{\frac{1}{2} - \bar{\omega}_2}{\bar{\gamma}_1} = \frac{\Delta_1}{2} + \kappa
	\label{eq:28}
\end{equation}
\begin{equation}
	\frac{\bar{\psi}_1}{2\bar{\gamma}_1} = \tilde{\psi}_1 = \psi_1 e^{i\Delta_1 \tau}; \quad \frac{\bar{\psi}_2}{2\bar{\gamma}_1} = \tilde{\psi}_2 = \psi_2 e^{i\Delta_2 \tau}
	\label{eq:29}
\end{equation}
\begin{equation}
	\frac{\partial \tilde{\psi}_1}{\partial \tau} = \frac{\partial \psi_1}{\partial \tau} e^{i\Delta_1 \tau} + i\Delta_1 \psi_1 e^{i\Delta_1 \tau}
	\label{eq:30}
\end{equation}
\begin{equation}
	\frac{\partial \tilde{\psi}_2}{\partial \tau} = \frac{\partial \psi_2}{\partial \tau} e^{i\Delta_2 \tau} + i\Delta_2 \psi_2 e^{i\Delta_2 \tau}
	\label{eq:31}
\end{equation}
In the limiting cases of $\Delta_1 \rightarrow 0$ and $\Delta_2 \rightarrow 0$, we obtain the following.

\begin{equation}
	\tilde{\psi}_1 = \psi_1; \quad \tilde{\psi}_2 = \psi_2
	\label{eq:32}
\end{equation}
\begin{equation}
	\frac{\partial \tilde{\psi}_1}{\partial \tau} = \frac{\partial \psi_1}{\partial \tau} + i\Delta_1 \psi_1
	\label{eq:33}
\end{equation}
\begin{equation}
	\frac{\partial \tilde{\psi}_2}{\partial \tau} = \frac{\partial \psi_2}{\partial \tau} + i\Delta_2 \psi_2
	\label{eq:34}
\end{equation}
Hence, we re-write Eq.~\ref{eq:25} and Eq.~\ref{eq:26} as
\begin{equation}
	\bm{
		\frac{\partial \psi_1}{\partial \tau}
		= -if - (1 + i\Delta_1)\psi_1 + i\psi_2^2
	}
	\label{eq:35}
\end{equation}
\begin{equation}
	\bm{
		\frac{\partial \psi_2}{\partial \tau}
		= -(\gamma_{21} + i\Delta_2)\psi_2 + 2i\psi_1 \psi_2^*
	}
	\label{eq:36}
\end{equation}

\section{Information processing capacity}
\label{sec:IPC}
We evaluated the information processing capacity \cite{dambre_information_2012_supp} of the reservoir. The decomposition required an Independent and Identically Distributed (i.i.d.) input as the basis is orthogonal only under that measure. Driving the reservoir with Mackey--Glass would violate the orthogonality the method depends on. We used an input $u(n) \sim \mathcal{U}[-1, 1]$, i.i.d. The encoder expects $s\in [0,1]$ so the drive is built from $(u+1)/2$ while the targets use $u$ directly.

Legendre polynomials $P_d$ are orthogonal on $[-1,1]$ under the uniform measure, and are normalized to zero mean and unit variance over the input. Basis functions are products over distinct delays,
\begin{equation}
	y_{\{d_i\}, \{\tau_i\}}(n) = \prod_i P_{d_i}\left( u \left(n-\tau_i\right)\right)
	\label{eq:37}
\end{equation}
enumerated with strictly increasing $\tau_i$ and degrees $d_i \ge 1$, so each combination of a delay set and a degree assignment appears exactly once. Caps of total degree $\sum_i d_i \le 11$, delay $\tau_i \le 25$, and at most 3 distinct delays per target give 447,161 basis functions.

\subsection{Capacity as a projection}
Let $Z \in \mathbb{R}^{T \times p}$ be the standardized feature matrix and $Q \in \mathbb{R}^{T\times r}$ an orthonormal basis for its column space, obtained by singular value decomposition. The capacity of a target $z$ is defined as the fraction of its variance a linear readout can reproduce, which is the $R^2$ of the regression of $z$ on the reservoir state.

Since the least-squares fitted values are the orthogonal projection of $z_c = z-\bar{z}$ onto the column space, with projector $\Pi=QQ^{\top}$,
\begin{equation}
	C\left[z\right] = \frac{\| \Pi z_c \|^2}{\| z_c \|^2} = \frac{\| Q^{\top} z_c \| ^2}{\| z_c \| ^2}.
	\label{eq:38}
\end{equation}
This form requires no matrix inversion and carries no regularization parameter, so the capacity measurement is independent of the choices examined in Sec.~\ref{sec:readout-reg}.

\subsection{The bias and its removal}
Consider a target $z$ statistically independent of the reservoir. After centering, $z_c$ lies in the $(T-1)$-dimensional subspace orthogonal to the constant vector, and is isotropically distributed within it. Its expected squared projection onto any fixed $r$-dimensional subspace of that space is the fraction $r/(T-1)$ of its squared norm, so from Eq.~\ref{eq:38},

\begin{equation}
	\mathbb{E}\left[C\right]_{\text{null}} = \frac{r}{T-1}.
	\label{eq:39}
\end{equation}

Raw capacities are therefore biased upward, and the bias is not negligible. At $r=256$ and $T=10^5$, it is $2.6 \times 10^{-3}$ per target, and the basis contains $\approx 4.5 \times 10^5$ targets. Since capacity is a coefficient of determination, the standard adjusted $R^2$ shrinkage applies,

\begin{equation}
	C_\text{adj} = 1 - \left(1 - C\right)\frac{T-1}{T-1-r}.
	\label{eq:40}
\end{equation}

Substituting Eq.~\ref{eq:39} verifies that this completely removes the bias.
\begin{equation}
	\mathbb{E}\big[C_\text{adj}\big]_{\text{null}}
	= 1 - \left(1 - \frac{r}{T-1}\right)\frac{T-1}{T-1-r}
	= 1 - \frac{T-1-r}{T-1}\cdot\frac{T-1}{T-1-r}
	= 0.
	\label{eq:41}
\end{equation}
This correction is applied to each target before thresholding rather than subtracted from the accumulated total, because thresholding is a nonlinear operation and the two orders do not commute.

\subsection{Rank truncation}
The noise floor of the estimator scales as $r/T$ by Eq.~\ref{eq:39}. At the full feature rank $r=2048$ with $T\approx 10^5$, this is $\approx 2 \times 10^{-2}$ per target, which accumulated over the basis exceeds the signal. We therefore restrict $Q$ to the leading $r=256$ principal directions, retaining $>99\%$ of the state variance and reducing the floor to $2.6 \times 10^{-3}$. The value 256 is taken from the number of components carrying 99\% of the variance at the richest operating point, Sec.~\ref{sec:echo-state-eff-dim}, so the discarded directions are those the readout barely uses.

\subsection{Significance threshold}
A surrogate null is constructed by taking real basis functions and randomly permuting them in time. This preserves each target's marginal distribution while destroying its temporal relation to the input. The threshold is the 99.9th percentile of the corrected capacities of 2000 such surrogates. Because Eq.~\ref{eq:41} centers the null on zero, this percentile measures genuine estimator scatter rather than a systemic offset. The measured surrogate mean is $-2.9 \times 10^{-8}$, which is negligible. Because the threshold is a percentile, one target in $10^{3}$ clears it on estimator noise alone, whatever the reservoir does. Retention counts are therefore reported relative to this false-positive expectation, since the basis grows combinatorially with degree and a high-degree cell could otherwise accumulate a large total from many individually marginal targets.

\subsection{Validation}
Dambre's theorem states that for a reservoir of rank $r$ driven by i.i.d. input, the total capacity summed over a complete orthogonal basis satisfies

\begin{equation}
	\sum_{\text{basis}} C \left[ y \right] \le r,
	\label{eq:42}
\end{equation}

with equality for a complete basis. A $K$-node linear delay line has exactly $K$ independent states and no nonlinearity, so its capacity must be exactly $K$, all at degree one, one unit per delay. For $K=20$ and $T=2 \times 10^4$, the estimator returns a total of 20.175 against a measured rank of 20 (with a relative error of 0.87\%). The degree one contribution was recovered as 20.000. The significance test reports degree two and four as consistent with noise.

\subsection{Results}
Medians over 20 independent startup realizations; comb entries carry their interquartile range, since both vary by $\approx 5$\%. Parametric resonance is identical in every realization.

\begin{table}[h]
	\centering
	\caption{Comparison of reservoir capacity across dynamical regimes. Medians over 20 independent startup realizations, with interquartile ranges given for the comb entries.}
	\alttext{Four rows comparing parametric resonance, the coherent comb and the chaotic comb: total information processing capacity, the fraction of the rank bound $r=256$ that this represents, capacity per available dimension, and the ratio to parametric resonance.}
	\label{tab:capacity_comparison}
	\begin{tabular}{lccc}
		\hline
		& \textbf{Parametric resonance} & \textbf{Coherent comb} & \textbf{Chaotic comb} \\
		\hline
		Total capacity
		& \textbf{160.32}
		& 18.0 (16.9--18.2)
		& 5.54 (5.49--5.82) \\

		Fraction of the bound $r=256$
		& \textbf{62.6\%}
		& 7.0\%
		& 2.2\% \\

		Capacity per available dimension
		& \textbf{0.626}
		& 0.070
		& 0.022 \\

		Ratio to parametric resonance
		& --
		& 8.9$\times$ lower
		& 29$\times$ lower \\
		\hline
	\end{tabular}
\end{table}

\begin{figure}[h]
	\centering
	\includegraphics[width=1\columnwidth]{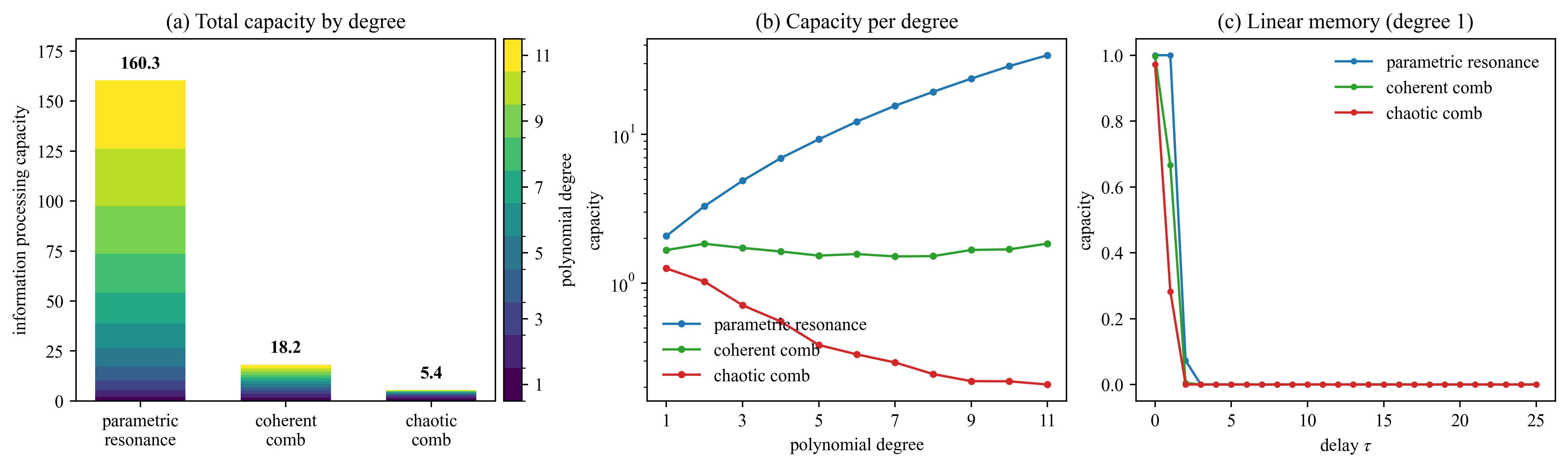}
	\caption{
	Capacity decomposition of parametric resonance, coherent comb and chaotic comb with $\text{F}_{\text{avg}}=40, 40, 20$ and $\Delta_1=-4.5, +4.5, +2.0$ respectively. The totals shown are for a single startup realization; Table~\ref{tab:capacity_comparison} reports medians over 20, which is why the comb totals differ slightly.
	}
	\alttext{Three panels. (a) A stacked bar chart of total information processing capacity for the three regimes, each bar segmented by polynomial degree from 1 to 11 on a colour scale. Parametric resonance reaches 160.3 with a visible contribution at every degree, while the coherent comb reaches 18.2 and the chaotic comb 5.4, both dominated by their lowest degrees. (b) Capacity per degree against polynomial degree on a logarithmic vertical axis: parametric resonance climbs steadily from about 2 at degree 1 to about 34 at degree 11, the coherent comb stays flat between about 1.5 and 1.9, and the chaotic comb falls from about 1.3 to 0.26. (c) Degree-one capacity against delay: all three regimes start near 1.0 at zero delay and fall to zero by delay 3, so their linear memory is short and closely comparable.}
\end{figure}
The regimes differ not only in the total capacity but also in where the capacity sits. At parametric resonance, the per-degree contribution rises monotonically to the cap and every degree is retained and the reservoir performs deep nonlinear transformation. The coherent comb neither deepens nor vanishes: its per-degree capacity stays flat between $1.5$ and $1.8$ at every degree. The chaotic comb decays, falling to within a factor of two of the false-positive expectation beyond degree three. Both comb regimes are shallower and lose exactly the high-degree nonlinear capability that one-step prediction of a chaotic series requires.

By Eq.~\ref{eq:42}, the totals are lower bounds, since any finite cap truncates the sum and the parametric resonance contribution is still rising at degree 11. The measurement establishes the ordering and the accessible depth and not a saturated total.

\section{Echo-state property and effective dimensionality}
\label{sec:echo-state-eff-dim}
\subsection{Effective dimensionality}

Let $\lambda_1 \ge \lambda_2 \ge \dots$ be the covariance eigenvalues of the standardized feature matrix. We use the participation ratio,
\begin{equation}
	D_{\text{eff}} = \frac{\left(\sum_i \lambda_i\right)^2}{\sum_i \lambda_i^2},
	\label{eq:43}
\end{equation}
which requires no arbitrary variance cutoff. If exactly $m$ eigenvalues equal $\lambda_0$ and the rest vanish,
\begin{equation}
	D_{\text{eff}} = \frac{\left(m \lambda_0\right)^2}{m \lambda_0^2} = m,
	\label{eq:44}
\end{equation}
so a state spread evenly over $m$ directions returns exactly $m$, and if one eigenvalue dominates, $D_{\text{eff}} \rightarrow 1$. By the Cauchy-Schwarz inequality applied to the $r$ non-zero eigenvalues, $\left( \sum_i \lambda_i \right)^2 \le r \sum_i \lambda_i^2$, so

\begin{equation}
	1 \le D_{\text{eff}} \le r,
	\label{eq:45}
\end{equation}
with the upper bound attained only when the spectrum is flat. $D_{\text{eff}}$ is therefore a continuous, threshold-free count of the directions actually carrying variance. Eigenvalues are then obtained from the singular values of the centered matrix as $\lambda_i=s_i^2/\left(T-1\right)$.

Eqs.~\ref{eq:35} and \ref{eq:36} are invariant under
\begin{equation}
	S: \quad \left(\psi_1, \psi_2\right) \longmapsto \left(\psi_1, -\psi_2\right),
	\label{eq:46}
\end{equation}
since the coupling enters Eq.~\ref{eq:35} as $i \psi_2^2$ with $i \left(- \psi_2\right)^2 = i \psi_2^2$ while every term of Eq.~\ref{eq:36} is odd in $\psi_2$ as the term $2i \psi_1 \psi_2^*$ changes sign with $\psi_2^*$. Every parametric state therefore has a symmetry partner.

Both terms on the right of Eq.~\ref{eq:36} carry a factor of $\psi_2$ or $\psi_2^*$, so $\psi_2=0$ is an invariant manifold.

\begin{equation}
	\psi_2=0 \implies \frac{\partial \psi_2}{\partial \tau} = - \left(
	\gamma_{21} + i \Delta_2
	\right) \cdot 0 + 2 i \psi_1 \cdot 0 = 0
	\label{eq:47}
\end{equation}

A trajectory that started exactly at 0 remains there. So for every simulation, we initialized it with a small random state rather than the origin as initializing at 0 would measure the trivial branch in place of the attractor.

Since the readout cannot distinguish the $S$ partners, both feature types, time and spectral, are invariant under $\psi_2 \rightarrow - \psi_2 : \lvert \psi_j \rvert ^2$ and $\lvert \mathcal{F} \left[\psi_j\right] \rvert$ by linearity of the transform. Two states related by $S$ produce identical feature matrices. This covariance-based dimensionality should not be interpreted as a count of spectral comb teeth. $D_{\text{eff}}$ measures the distribution of state variance across principal directions, whereas the number of comb teeth counts resolved frequency components. The two quantities can therefore vary independently.

\subsection{Measuring convergence}
We drive $M=12$ replicas from different initial conditions with an identical input sequence and track the mean pairwise distance,

\begin{equation}
	d_{\text{raw}}(n) = \langle
	\| \psi^{(i)} (n) - \psi^{(j)} (n) \|
	\rangle_{i<j},
	\label{eq:48}
\end{equation}
\begin{equation}
	d_{\text{sym}}(n) = \langle \min \left( \| \psi^{(i)} - \psi^{(j)} \|, \| \psi^{(i)} - S \psi^{(j)} \| \right) \rangle _{i<j}.
	\label{eq:49}
\end{equation}
Due to the invariance just established, $d_{\text{sym}}$, the distance on the quotient space $\mathbb{R}^4/S$ is the quantity relevant to computation. Using $d_{\text{raw}}$ alone reports a failure of the echo-state property at parametric resonance, where replicas divide between the two branches while the quotient distance falls to $\sim 2 \times 10^{-15}$ and the reservoir has converged in every respect the readout can perceive. Perturbations are drawn at the amplitude scale of the driven attractor at each operating point and the convergence rates are obtained by least squares on
\begin{equation}
	\ln d(n) = \ln d(0) - \lambda n
	\label{eq:50}
\end{equation}
over the region preceding the numerical floor, with the coefficient of determination reported.
\begin{table}[h]
	\centering
	\caption{Dynamical dimensionality, feature activation, and convergence characteristics across operating regimes.}
	\alttext{Six rows across the four operating regimes: effective dimensionality, the number of principal components needed for 95\% and 99\% of the variance, the variance fraction carried by the leading component, the number of active features out of 2048, the fitted convergence rate per symbol, and the coefficient of determination of that fit.}
	\label{tab:dynamical_dimensionality}
	\begin{tabular}{lcccc}
		\hline
		& \textbf{Sub-threshold}
		& \textbf{Parametric resonance}
		& \textbf{Coherent comb}
		& \textbf{Chaotic comb} \\
		\hline
		$D_\text{eff}$
		& 1.91
		& 2.77
		& \textbf{7.15}
		& \textbf{7.40} \\

		Components for 95\% / 99\% variance
		& 2 / 2
		& 8 / 11
		& 30 / 89
		& 54 / 179 \\

		Leading component, variance fraction
		& 0.607
		& 0.565
		& 0.310
		& 0.307 \\

		Active features (of 2048)
		& \textbf{1024}
		& 2048
		& 2048
		& 2048 \\

		Convergence rate $\lambda$ (per symbol)
		& 4.99
		& \textbf{4.03}
		& $2.5\times10^{-4}$
		& $2.1\times10^{-4}$ \\

		Exponential fit $R^{2}$
		& 1.000
		& 0.974
		& 0.011
		& 0.004 \\
		\hline
	\end{tabular}
\end{table}

\begin{figure}[h]
	\centering
	\includegraphics[width=1\columnwidth]{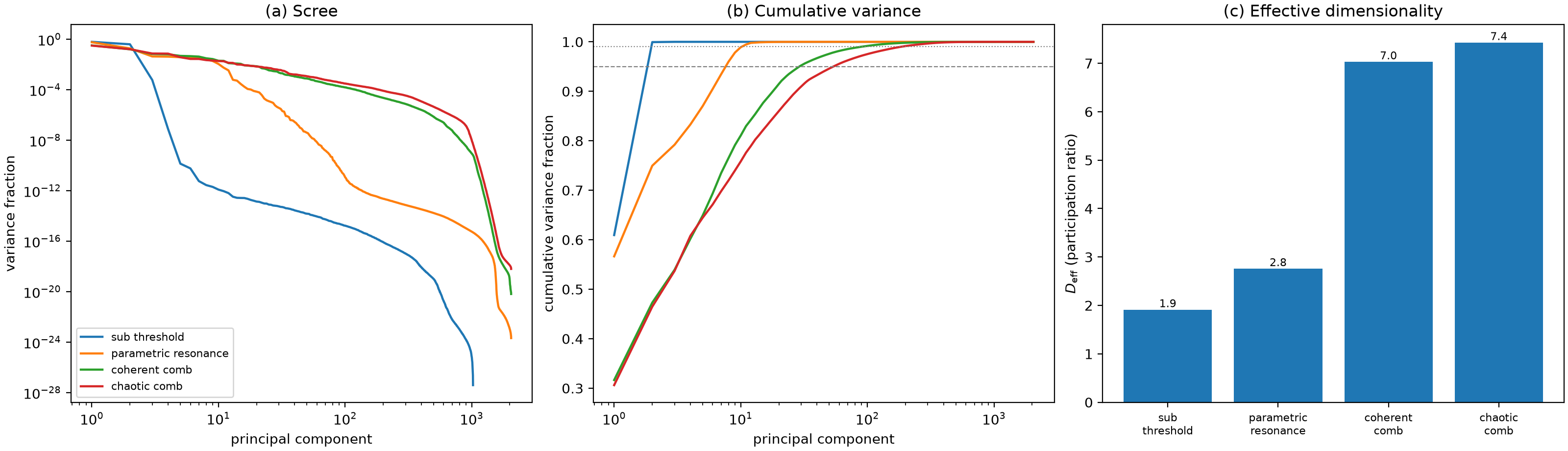}
	\caption{(a) Variance fraction per principal component. Sub-threshold reaches the numerical floor by component $\approx 5$, so its long tail below $10^{-16}$ is floating-point noise rather than structure. (b) Cumulative variance with 95\% and 99\% guides. (c) Participation ratio, Eq.~\ref{eq:43}.}
	\alttext{Three panels. (a) Scree plot of variance fraction against principal-component index, both axes logarithmic, for the four regimes. The sub-threshold curve falls steeply and reaches the numerical floor near $10^{-28}$; the two comb curves decay slowly and remain highest across the whole range, with parametric resonance between them. (b) Cumulative variance fraction against component index, with dashed and dotted guides at 95\% and 99\%. Sub-threshold saturates within two components and parametric resonance within about eleven, while the comb curves need tens to hundreds. (c) Bar chart of the participation ratio: 1.9 sub-threshold, 2.8 parametric resonance, 7.0 coherent comb and 7.4 chaotic comb.}
\end{figure}
\begin{figure}[h]
	\centering
	\includegraphics[width=1\columnwidth]{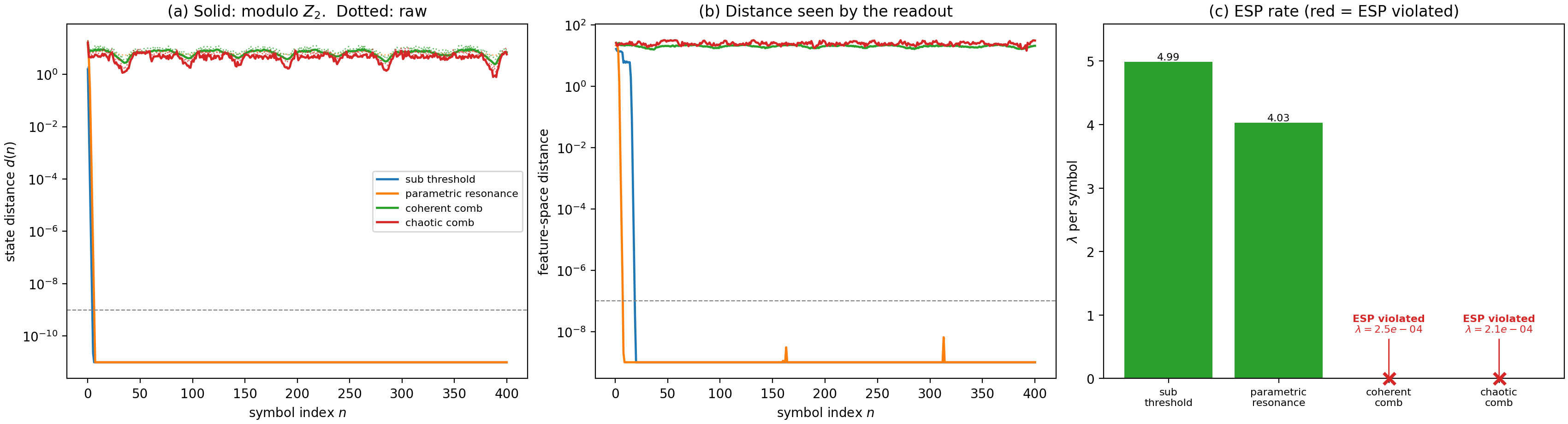}
	\caption{Mean pairwise distance between 12 replicas driven by identical input from different initial conditions. Parametric resonance contracts to machine precision; neither comb regime converges.}
	\alttext{Three panels. (a) Mean pairwise state distance between replicas against symbol index on a logarithmic vertical axis, with solid curves for the distance modulo the $Z_2$ symmetry and dotted curves for the raw distance. Sub-threshold and parametric resonance collapse from about 10 to the $10^{-11}$ floor within a few symbols, while both comb curves stay near 10 for all 400 symbols. (b) The same comparison measured in the log-compressed feature space the readout actually sees, showing the same separation. (c) Bar chart of the fitted convergence rate per symbol. Sub-threshold and parametric resonance appear as tall green bars at 4.99 and 4.03. Both comb regimes have rates of order $10^{-4}$, too small to draw at this scale, and are instead marked with a red cross on the baseline and labelled \emph{ESP violated} together with their fitted values of $2.5\times10^{-4}$ and $2.1\times10^{-4}$.}
\end{figure}
Effective dimensionality rises monotonically into the comb regimes, $1.91 \to 2.77 \to 7.15 \to 7.40$, while the echo-state property collapses: parametric resonance contracts at 4.03 per symbol with a clean exponential fit ($R^2 = 0.974$). Both comb regions have fitted rates indistinguishable from zero with $R^2<0.02$. The exponential model does not describe them as there is no decay to describe.

Below threshold, $\psi_2 \equiv 0$, so the 1024 mode-2 columns are constant and dropped, leaving 1024 active features. The low $D_{\text{eff}}$ is not a consequence of this halving, since constant columns carry no variance and do not contribute to the participation ratio; rather, the surviving mode-1 features are effectively two-dimensional, with the first two components accounting for 60.67\% and 39.27\% of the variance (99.94\% combined), yielding $D_{\text{eff}} = 1.91$.

Between parametric resonance and coherent comb, $D_{\text{eff}}$ rises by $2.5\times$ whereas the count of components reaching 99\% of the variance rises by $8\times$. As Eq.~\ref{eq:43} is dominated by the largest eigenvalues, the 99\% count reaches into the tail. This disparity localizes the comb regimes' extra dimensionality in weakly excited directions. These directions are suppressed by the readout penalty, and are too weak to be estimated reliably.

\section{Composition of feature vector}
\label{sec:feature-vec}
The block index sets for the feature vector are:
\begin{equation}
	\mathcal{T} = \left\{1, \dots, 1024\right\}, \quad \mathcal{S} = \left\{ 1025, \dots, 2048 \right\}.
	\label{eq:51}
\end{equation}

The time block $\mathcal{T}$ contains $| \psi_j \left( \tau_k \right) |^2$ and therefore discards the phase of $\psi_j$ entirely. The spectral block $\mathcal{S}$ contains $\vert \mathcal{F} \left[ \psi_j \right] \vert$, obtained from the complex modal amplitudes, and is therefore sensitive to intra-symbol phase evolution and phase relationships between harmonic components that are absent from the power trajectories. Applying the Fourier transform to the sampled power would retain only information already represented by the power-based temporal block and would substantially increase redundancy between the two feature blocks. The use of complex amplitudes therefore provides a complementary phase-sensitive representation.

\subsection{Ablation}
Comparing readouts trained on column subsets of a single reservoir realization ensures that the trajectories, warm-up, and train/test split are identical across all arms, with only the readout input differing. Each arm uses its own cross-validated penalty, Sec.~\ref{sec:readout-reg}, since the optimal penalty depends on dimensionality and a common value would mix regularization effects with information content.
\begin{table}[h]
	\centering
	\caption{Prediction error for temporal and spectral feature arms across dynamical regimes.}
	\alttext{Test NMSE for three readout feature sets, time-domain block only, spectral block only, and both combined, across the four operating regimes.}
	\label{tab:feature_arm_comparison}
	\begin{tabular}{lcccc}
		\hline
		& \textbf{Sub-threshold}
		& \textbf{Parametric resonance}
		& \textbf{Coherent comb}
		& \textbf{Chaotic comb} \\
		\hline
		Time only ($\mathcal{T}$)
		& $1.98\times10^{-2}$
		& $3.31\times10^{-3}$
		& $1.03\times10^{-2}$
		& $1.14\times10^{-1}$ \\

		Spectral only ($\mathcal{S}$)
		& $4.09\times10^{-2}$
		& $2.04\times10^{-3}$
		& $2.64\times10^{-2}$
		& $1.25\times10^{-1}$ \\

		\textbf{Both}
		& $\mathbf{1.95\times10^{-2}}$
		& $\mathbf{4.51\times10^{-4}}$
		& $\mathbf{1.26\times10^{-2}}$
		& $\mathbf{6.39\times10^{-2}}$ \\
		\hline
	\end{tabular}
\end{table}
At parametric resonance, the combination outperforms the time block by $7.35 \times$, and the spectral block by $4.5\times$, which is possible only when each block carries different information. These values were validated against 20 experiments.

\subsection{Rank analysis}
The subspace overlap is quantified by the principal angles between the two blocks. Let $Q_a, Q_b \in \mathbb{R}^{T \times k}$ be the orthonormal bases for the leading $k=20$ principal directions of the standardized blocks $\mathcal{T}$ and $\mathcal{S}$. The principal angles $0 \le \theta_1 \le \dots \le \theta_k \le \pi / 2 $ are defined by
\begin{equation}
	\cos\theta_1 = \max_{a \in \text{col}\,Q_a,\ b \in \text{col}\,Q_b} \frac{\left|a^{\top}b\right|}{\lVert a\rVert\,\lVert b\rVert}
	\label{eq:52}
\end{equation}
with subsequent angles maximizing over directions orthogonal to those already chosen. Equivalently, the cosines are the singular values of the cross-product of the bases,
\begin{equation}
	\cos \theta_k = \sigma_k \left(
	Q_a^{\top}Q_b
	\right) \in [0, 1].
	\label{eq:53}
\end{equation}
The value of 1 indicates a direction lying in both subspaces, i.e., duplicate information, and a value of 0 indicates a direction in one subspace orthogonal to the whole of the other. If $\mathcal{S}$ was a re-expression of $\mathcal{T}$, then all $\cos\theta_k=1$.

\begin{figure}[h]
	\centering
	\includegraphics[width=1\columnwidth]{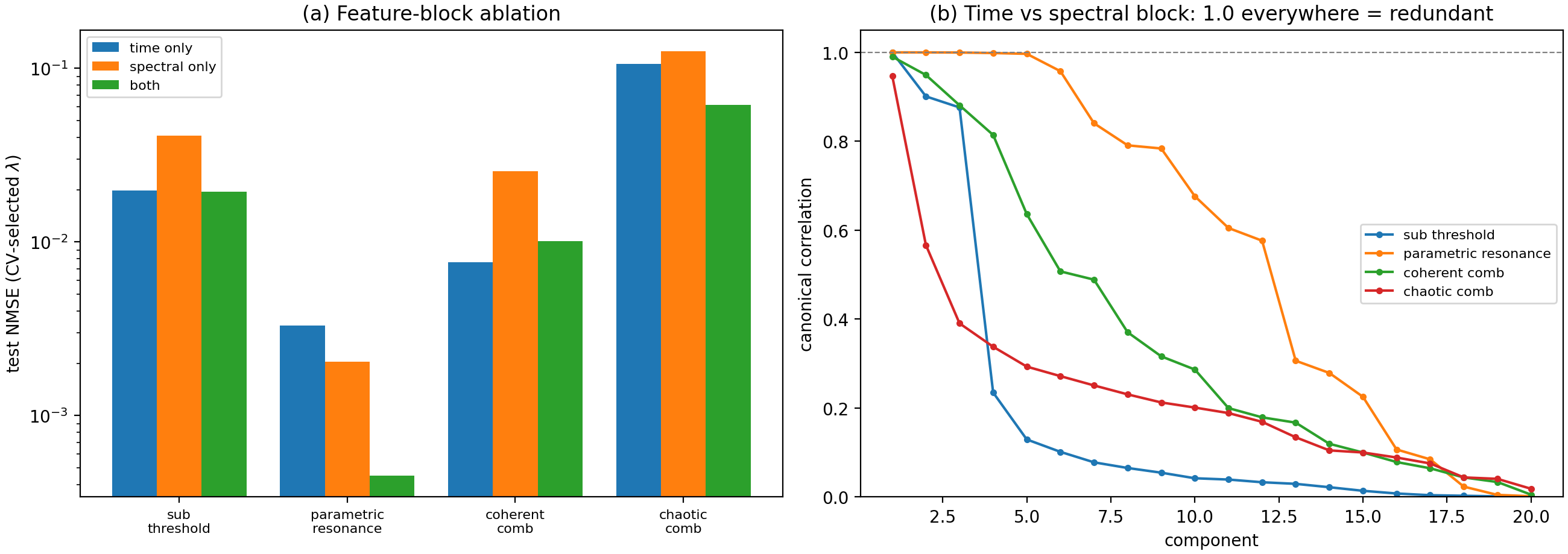}
	\caption{(a) Test error for readouts restricted to block subsets, each at its own cross-validated penalty. (b) Canonical correlations, Eq.~\ref{eq:53}. A value of 1.0 across all components would indicate the blocks span the same subspace.}
	\alttext{Two panels. (a) Grouped bar chart of test NMSE on a logarithmic axis for the four regimes, with three bars each for the time-only, spectral-only and combined feature blocks. At parametric resonance the combined bar sits far below either single block, whereas at sub-threshold it is indistinguishable from the time-only bar. (b) Canonical correlation against component index from 1 to 20 for the four regimes, with a dashed line at 1.0 marking complete redundancy. Parametric resonance stays near 1.0 for the first five components before decaying to near zero by component 20, while sub-threshold falls fastest, dropping below 0.25 by component 4.}
\end{figure}
At parametric resonance the canonical correlations run from a maximum of 1.000 through a median of 0.641 to a minimum of 0.001. The subspaces overlap substantially but are far from coincident. The leading correlation is near unity in every regime and carries no information, since both blocks encode the same total energy per symbol; only the decay is informative. A low canonical correlation arises either when the blocks carry genuinely different signal or when one block is dominated by noise, since independent noise is orthogonal to everything. Sub-threshold is the second case: $\psi_2 \equiv 0$ renders half of each block inert; its median correlation is 0.040 — the lowest of the four regimes — yet combining the blocks yields nothing ($1.02\times$). Eq.~\ref{eq:53} can therefore refute the hypothesis that the blocks span the same subspace, but cannot on its own establish that their difference is useful. The ablation and the angle analysis must be read together, and together they show the spectral block earning its place precisely where the reservoir operates.

\section{Coherent and chaotic comb}
\label{sec:coherent-chaotic}
We adopt the largest Lyapunov exponent, since it classifies the dynamics directly rather than through a spectral proxy.

\subsection{Estimator}
For the system holding the drive constant, the largest Lyapunov exponent is
\begin{equation}
	\lambda_{\max} = \lim_{N \rightarrow \infty} \frac{1}{N T_r} \sum_{k=1}^{N} \ln \frac{\| \delta_k \|}{d_0}
	\label{eq:54}
\end{equation}
computed by the algorithm of Benettin \cite{benettin_kolmogorov_1976_supp, benettin_lyapunov_1980_supp}. Two trajectories initially separated by $d_0=10^{-7}$ are evolved for an interval $T_r$. The separation $\delta_k$ at the end of the interval $k$ is recorded, and the perturbed trajectory is then rescaled back to $d_0$ along with the current separation direction before the next interval.

We chose $d_0=10^{-7}$ as it sits well above the integration tolerance of $10^{-10}$, so the measurement is not dominated by numerical error, and far below the attractor scale, so linearization remains valid. Each trajectory pair is initialized off the invariant manifold that Eq.~\ref{eq:47} identifies.

\subsection{Classification and thresholds}
\begin{equation}
	\text{chaotic if } \lambda_{\max} > 10^{-2}; \quad \text{sub-threshold if } |\psi_2| < 10^{-6}.
	\label{eq:55}
\end{equation}
Chaos and the comb are independent properties, so a second, purely analytic condition is required. We use the detuning boundary of Qi \emph{et al.} \cite{qi_existence_2020_supp},
\begin{equation}
	2\Delta_1\Delta_2 \le -\left(1 + \Delta_1^2 + 2\gamma_{21}\right),
	\label{eq:comb-condition}
\end{equation}
evaluated over the drive excursion. Eqs.~\ref{eq:55} and~\ref{eq:comb-condition} together give the five levels of the regime map: sub-threshold; parametric resonance, above threshold with Eq.~\ref{eq:comb-condition} not satisfied and $\lambda_{\max} \le 10^{-2}$; coherent comb, with Eq.~\ref{eq:comb-condition} satisfied and $\lambda_{\max} \le 10^{-2}$; chaotic comb, with both Eq.~\ref{eq:comb-condition} satisfied and $\lambda_{\max} > 10^{-2}$; and chaotic non-comb, with $\lambda_{\max} > 10^{-2}$ but Eq.~\ref{eq:comb-condition} not satisfied.
For a limit cycle, a perturbation along the direction of the flow merely translates the state along the orbit, producing neither growth nor decay. For a fixed point, all directions contract and $\lambda_{\max} \approx -\gamma_1=-1$ in normalized units. Finite-time estimates of a true zero scatter by $\sim 10^{-3}$. The value $10^{-2}$ therefore sits an order of magnitude above the scatter and two orders below the fixed-point exponent, separating all three cases cleanly.

\subsubsection{Threshold robustness}
The threshold $\lambda_{\max}>10^{-2}$ is set relative to the finite-time scatter of the Benettin estimator. Taking the standard deviation of the running estimate over the final quarter of each run as a per-cell noise measure gives a median scatter of $2.3\times10^{-4}$ and a 99th percentile of $7.3\times10^{-3}$ across the excited grid. The adopted threshold therefore lies above the observed estimator scatter for near-zero exponents.

The classification is insensitive to the precise threshold. Within the comb region, the chaotic comb occupies 45.9\%, 45.8\%, 45.7\%, and 45.4\% of cells for thresholds of $3\times10^{-3}$, $10^{-2}$, $3\times10^{-2}$, and $10^{-1}$, respectively. The classification changes appreciably only when the threshold is reduced to $10^{-3}$ or below, where it enters the estimator scatter band. The operating points used in the manuscript are also well separated: the coherent comb has $\lambda_{\max}=3\times10^{-5}$, whereas the chaotic comb has $\lambda_{\max}=0.563$.

As an independent check, the normalized spectral entropy of the mode-2 power trace gives the same ordering of the three excited regimes, with values of 0.015, 0.108, and 0.631 for parametric resonance, coherent comb, and chaotic comb, respectively. The spectral entropy is used only as an independent diagnostic and not as the regime classification criterion.

\subsubsection{Second metric}
We also compute the normalized spectral entropy of the mode-2 power trace,
\begin{equation}
	H = - \frac{1}{\ln N} \sum_{i} p_i \ln p_i, \quad p_i = \frac{
		\lvert \mathcal{F} \left[ | \psi_2 | ^2 - \langle | \psi_2 |^2 \rangle \right]_i \rvert^2
	}{\sum_{j} \lvert \cdot \rvert^2_j}
	\label{eq:56}
\end{equation}
with the mean subtracted so the zero-frequency bin does not dominate the normalization. $H=0$ places all power in one spectral line and $H=1$ is a flat broadband spectrum.
\begin{table}[h]
	\centering
	\caption{Dynamical indicators across the four operating regimes.}
	\alttext{Four rows across the four operating regimes: the largest Lyapunov exponent per unit normalized time and per symbol, the normalized spectral entropy of the mode-2 power trace, and the mode-2 amplitude.}
	\label{tab:lyapunov_entropy}
	\begin{tabular}{lcccc}
		\hline
		& $\lambda_\text{max}$ (per $\tau$)
		& $\lambda_\text{max}$ (per symbol)
		& $H$
		& $\lvert\psi_2\rvert$ \\
		\hline
		Sub-threshold
		& $-1.0003$
		& $-5.002$
		& $0.363$
		& $7\times10^{-10}$ \\

		Parametric resonance
		& $-0.0066$
		& $-0.033$
		& $\mathbf{0.015}$
		& $8.002$ \\

		Coherent comb
		& $+0.00003$
		& $+0.0002$
		& $0.108$
		& $3.594$ \\

		Chaotic comb
		& $\mathbf{+0.5627}$
		& $\mathbf{+2.814}$
		& $\mathbf{0.631}$
		& $10.686$ \\
		\hline
	\end{tabular}
\end{table}
The sub-threshold exponent recovers $-\gamma_1 = -1$ to three decimal places, which validates the estimator against a case whose answer is known analytically. Only the chaotic comb has a positive exponent. Across the three excited regimes the two metrics order the dynamics identically despite sharing no computation; sub-threshold is the exception, where $|\psi_2|\sim10^{-9}$ leaves the entropy measuring numerical noise rather than the dynamics.

\begin{figure}
	\centering
	\includegraphics[width=1\columnwidth]{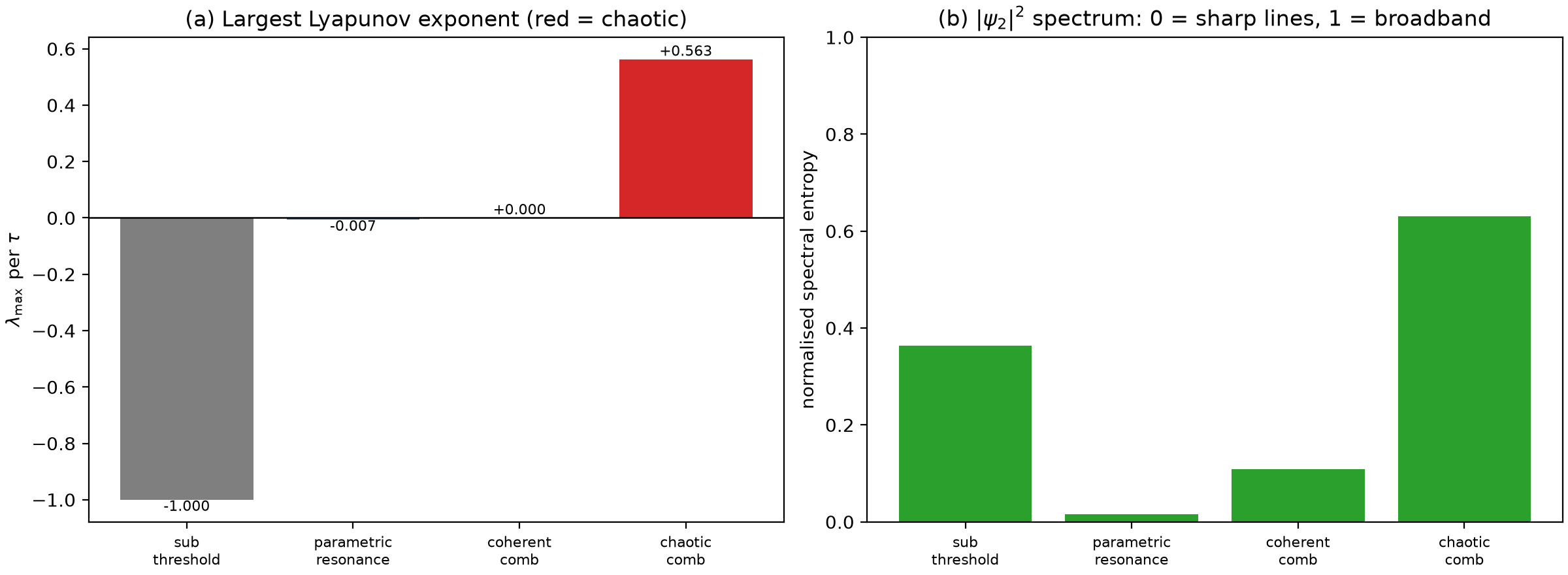}
	\caption{(a) Largest Lyapunov exponent per $\tau$ at the four operating points, computed at constant drive by the Benettin algorithm; the chaotic comb carries the only positive value. (b) Normalized spectral entropy of the mode-2 power trace, Eq.~\ref{eq:56}, at the same four points. The two metrics order the three excited regimes identically.}
	\alttext{Two bar charts. (a) The largest Lyapunov exponent per unit normalized time at the four operating points: $-1.000$ sub-threshold, $-0.007$ parametric resonance, $+0.000$ coherent comb and $+0.563$ chaotic comb, the last highlighted in red as the only positive value. (b) Normalized spectral entropy of the mode-2 power trace: 0.36 sub-threshold, 0.02 parametric resonance, 0.11 coherent comb and 0.63 chaotic comb.}
	\label{fig:chao-vs-coh-2-metrics}
\end{figure}

\begin{figure}
	\centering
	\includegraphics[width=1\columnwidth]{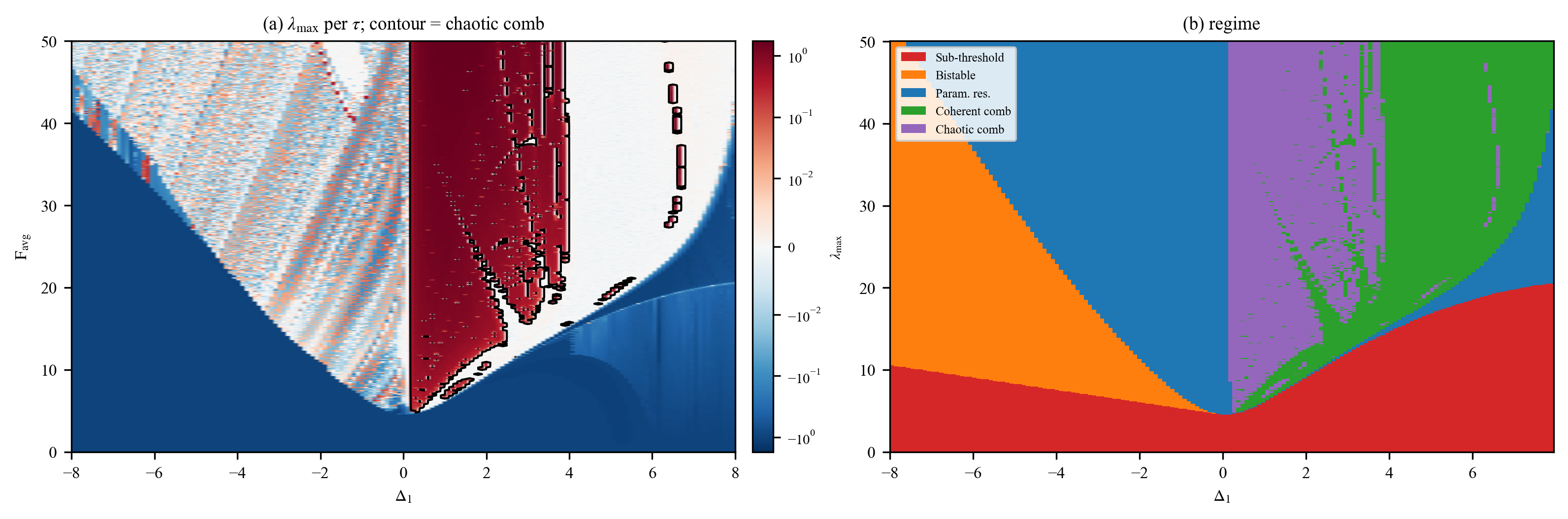}
	\caption{Regime structure over the $(\text{F}_\text{avg},\Delta_1)$ plane at the resolution of the parameter sweeps, $503\times161$ cells. (a) Largest Lyapunov exponent, one evaluation of Eq.~\ref{eq:54} per cell. The color scale is symmetric-logarithmic about zero: $\lambda_\text{max}$ spans $-1.07$ to $+1.73$, while the coherent/chaotic distinction lies within $\pm10^{-2}$ of zero and a linear scale would render it as a single flat color. The black contour is the chaotic-comb boundary, $\lambda_\text{max} = 10^{-2}$ restricted to cells satisfying the comb condition. (b) The resulting five-level regime map.}
	\alttext{Two maps over the plane of average drive amplitude against detuning. (a) The largest Lyapunov exponent on a symmetric-logarithmic blue-to-red scale: deep blue where the dynamics contract strongly, near-white where the exponent is close to zero, and dark red in a tongue at positive detuning between roughly 0.2 and 4. A black contour outlines the chaotic comb; fine striations run through the parametric-resonance region and isolated islands appear near detuning 6.5 at high drive. (b) The resulting five-level regime map, with sub-threshold red at low drive, bistable orange, parametric resonance blue at negative detuning, and coherent comb green and chaotic comb purple at positive detuning.}
\end{figure}

Of the 80,983 cells, 36.14\% are sub-threshold, 41.35\% coherent, and 22.51\% exceed the chaos threshold. Chaos and the comb condition are independent properties, and separating them matters: 16.21\% of all cells also satisfy the comb condition and constitute the chaotic comb, occupying a connected tongue at $\Delta_1\in[0.2,6.7]$ with $\lambda_{\max}$ from $0.011$ to $1.731$. The remaining 6.30\% are chaotic but not comb and are far weaker, $\lambda_{\max}$ reaching only $0.549$; the contour is restricted to the comb region so that it delimits the chaotic comb rather than chaos in general.

Fine striations through the parametric-resonance region where $\lambda_\text{max}$ hovers about zero, and isolated chaotic islands near $\Delta_1\approx6.5$ at high drive were also observed. The non-comb population sits close to the classification threshold and is excluded from the chaotic-comb count and from the boundary contour, since neither the comb condition nor a converged positive exponent is established for it.

\section{Branch selection in the bistable region}
\label{sec:branch-sel}
\subsection{Stability of the trivial branch}
The bistable region is bounded by two drive amplitudes, and the upper one follows from linear stability analysis. On the trivial branch $\psi_2=0$, the steady state of Eq.~\ref{eq:35} is:

\begin{equation}
	0 = -if - \left(1+i\Delta_1\right)\psi_1^{0}
	\quad\Longrightarrow\quad
	\psi_1^{0} = \frac{-if}{1+i\Delta_1},
	\qquad
	\big|\psi_1^{0}\big|^{2} = \frac{f^{2}}{1+\Delta_1^{2}}.
	\label{eq:57}
\end{equation}
Perturbing $\psi_2 = \epsilon$ and linearizing Eq.~\ref{eq:36} about this state gives a system coupling $\epsilon$ to its own conjugate — the signature of a parametric instability:

\begin{equation}
	\frac{\partial}{\partial\tau}
	\begin{pmatrix} \epsilon \\ \epsilon^{*} \end{pmatrix}
	=
	\begin{pmatrix}
		-\left(\gamma_{21}+i\Delta_2\right) & 2i\psi_1^{0} \\
		-2i\big(\psi_1^{0}\big)^{*} & -\left(\gamma_{21}-i\Delta_2\right)
	\end{pmatrix}
	\begin{pmatrix} \epsilon \\ \epsilon^{*} \end{pmatrix} .
	\label{eq:58}
\end{equation}
The characteristic equation is:
\begin{equation}
	\left(s+\gamma_{21}+i\Delta_2\right)\left(s+\gamma_{21}-i\Delta_2\right) - 4\big|\psi_1^{0}\big|^{2} = 0\quad\Longrightarrow\quad\left(s+\gamma_{21}\right)^{2} = 4\big|\psi_1^{0}\big|^{2} - \Delta_2^{2} ,
	\label{eq:59}
\end{equation}
so the growth rates are
\begin{equation}
s_{\pm} = -\gamma_{21} \pm \sqrt{4\big|\psi_1^{0}\big|^{2} - \Delta_2^{2}} .
\label{eq:60}
\end{equation}
The trivial branch loses stability when $\text{Re}\,s_+ > 0$, which requires the radical to be real and exceed $\gamma_{21}$:
\begin{equation}
	4\big|\psi_1^{0}\big|^{2} - \Delta_2^{2} > \gamma_{21}^{2}
	\;\Longleftrightarrow\;
	\frac{4f^{2}}{1+\Delta_1^{2}} > \gamma_{21}^{2}+\Delta_2^{2}
	\;\Longleftrightarrow\;
	f > \tfrac12\sqrt{\left(1+\Delta_1^{2}\right)\left(\gamma_{21}^{2}+\Delta_2^{2}\right)} \equiv f_\text{upper} .
	\label{eq:61}
\end{equation}
This identifies $f_\text{upper}$ as the drive above which the trivial state is unstable, so bistability ends there and the parametric branch is the only attractor. The lower boundary, at which the parametric branch first comes into existence, is the Arnold-tongue threshold
\begin{equation}
	f_\text{Arnold} = \tfrac12\left|\gamma_{21}\Delta_1 + \Delta_2\right| ,
	\label{eq:62}
\end{equation}
and for $f_\text{Arnold} < f < f_\text{upper}$ both states are stable and the realized branch is decided by the startup rather than by the parameters.

As $\Delta\text{F}$ modulates the drive, the reservoir visits the whole excursion $f \in [\tfrac12\text{F}_\text{avg}, \tfrac32\text{F}_\text{avg}]$ within every symbol. Genuine bistability requires the entire excursion to lie inside the band, i.e., $\tfrac12\text{F}_\text{avg} > f_\text{Arnold}$ and $\tfrac32\text{F}_\text{avg} < f_\text{upper}$. At $\Delta_1 = -4.5$, where Eqs.~\ref{eq:61} and~\ref{eq:62} give $f_\text{Arnold} = 7.875$ and $f_\text{upper} = 26.03$, this restricts
\begin{equation}
	15.75 \le \text{F}_\text{avg} \le 17.35 ,
	\label{eq:63}
\end{equation}
considerably narrower than the thresholds themselves suggest.

\subsection{Sensitivity analysis}
Comparing the two startup protocols, we use Monte Carlo sampling over 32 initial conditions at each of four amplitude scales $\sigma \in \{0.01, 0.1, 1, 8\}$ and across 24 values of $\text{F}_\text{avg}$. The first protocol holds the drive at $\text{F}_\text{avg}$ for 50 symbols. The second starts above $f_\text{upper}$, where Eq.~\ref{eq:61} guarantees that the trivial branch is unstable, and then ramps adiabatically down to $\text{F}_\text{avg}$. This removes initialization dependence by construction. The branch is identified from the final $\lvert\psi_2\rvert$.

\begin{figure}
	\centering
	\includegraphics[width=1\columnwidth]{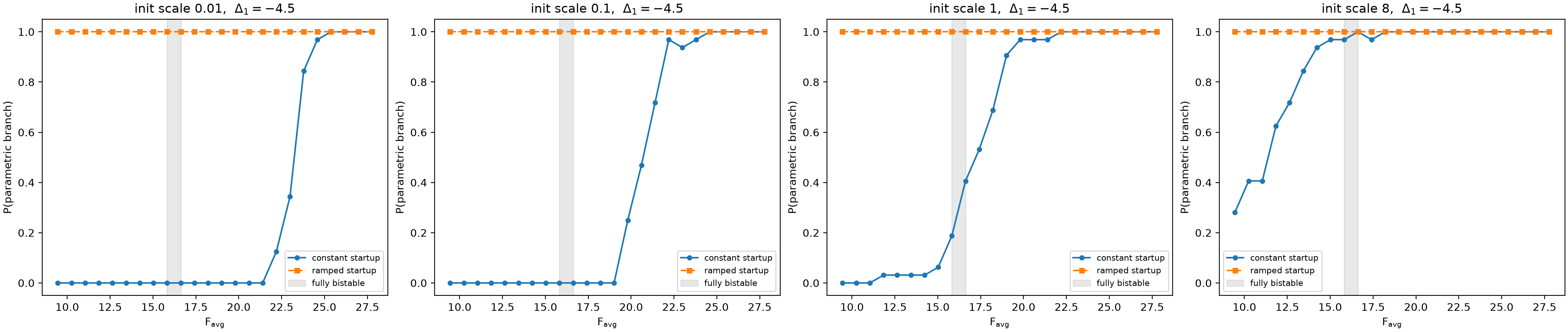}
	\caption{Probability of selecting the parametric branch against $\text{F}_\text{avg}$ at $\Delta_1 = -4.5$, for four initial-condition scales. The ramped protocol reaches the parametric branch everywhere; the constant protocol crosses over at an amplitude that depends strongly on the initial condition. Shading marks the band of Eq.~\ref{eq:63}.}
	\alttext{Four panels, one for each initial-condition amplitude scale of 0.01, 0.1, 1 and 8. Each plots the probability of selecting the parametric branch against average drive amplitude, with a blue curve for the constant startup, a flat orange line at 1.0 for the ramped startup, and a grey band marking the fully bistable range. The blue curve is a sigmoid whose midpoint shifts to lower drive as the initial scale grows, from about 23.4 at scale 0.01 to about 11.5 at scale 8, while the ramped result stays at 1.0 throughout.}
	\label{fig:branch-selection}
\end{figure}
Within that band, at the amplitude scale the solver actually uses ($\sigma = 0.01$), the constant protocol selects the parametric branch in 0\% of trials and the ramped protocol in 100\%. The crossover is a basin-of-attraction effect rather than a threshold effect: the constant protocol reaches 50\% selection at
\begin{table}[h]
	\centering
	\begin{tabular}{lcccc}
		\hline
		Initial scale $\sigma$
		& 0.01
		& 0.1
		& 1
		& 8 \\
		\hline
		$\text{F}_\text{avg}$ at 50\% selection
		& 23.4
		& 20.7
		& 17.3
		& 11.5 \\
		\hline
	\end{tabular}
\end{table}
Above $\text{F}_\text{avg} \approx 24.6$, the two protocols agree. This is slightly below the analytical threshold $f_\text{upper}=26.03$ in Eq.~\ref{eq:61}, since the constant-drive startup approaches the loss of stability of the trivial branch as $f$ approaches $f_\text{upper}$. Thus, a 50-symbol startup can leave the trivial branch slightly before the analytical threshold.

Within the bistable band the reservoir therefore occupies the trivial branch, where $\psi_2 \equiv 0$ leaves half the feature vector inert, so that region is closer to sub-threshold operation than to parametric resonance. Operating points whose excursion carries them above $f_\text{upper}$ are unaffected: the parametric-resonance point ($\text{F}_\text{avg} = 40$, peak drive 60 against $f_\text{upper} = 26.03$) is driven onto the parametric branch within each symbol and both protocols settle at $\lvert\psi_2\rvert = 8.002$, agreeing to five significant figures over 32 initial conditions and all four scales. The ramped protocol removes the ambiguity and is adopted.

\section{Readout regularization}
\label{sec:readout-reg}
The readout minimizes the penalized residual sum of squares,
\begin{equation}
	\boldsymbol{\beta}(\lambda) = \arg\min_{\boldsymbol{\beta}} \Big\{ \big\lVert \mathbf{y} - Z\boldsymbol{\beta} \big\rVert^{2} + \lambda\big\lVert\boldsymbol{\beta}\big\rVert^{2} \Big\} .
	\label{eq:64}
\end{equation}
Substituting the singular value decomposition $Z = USV^{\top}$ into the normal equations gives
\begin{equation}
	\boldsymbol{\beta}(\lambda) = \sum_i \mathbf{v}_i \frac{s_i}{s_i^{2}+\lambda}\,\mathbf{u}_i^{\top}\mathbf{y} ,
	\label{eq:65}
\end{equation}
so a single factorization serves every $\lambda$, which is what makes an exhaustive search affordable.

The fitted values are $\hat{\mathbf{y}} = Z\boldsymbol{\beta} = H(\lambda)\,\mathbf{y}$ with hat matrix
\begin{equation}
	H(\lambda) = U\,\text{diag}\!\left(\frac{s_i^{2}}{s_i^{2}+\lambda}\right)U^{\top} ,
	\label{eq:66}
\end{equation}
so the effective number of parameters — the trace of the hat matrix, which for ordinary least squares equals the parameter count — is
\begin{equation}
	\nu(\lambda) = \text{tr}\,H(\lambda) = \sum_i \frac{s_i^{2}}{s_i^{2}+\lambda} .
	\label{eq:67}
\end{equation}
Eqs.~\ref{eq:65},~\ref{eq:66},~\ref{eq:67} identify $\lambda$ as a soft threshold on the singular value spectrum; directions with $s_i^{2}\gg\lambda$ pass unattenuated and count fully towards $\nu$, those with $s_i^{2}\ll\lambda$ are suppressed and count for nothing, with $\nu\to r$ as $\lambda\to0$ and $\nu\to0$ as $\lambda\to\infty$. When the feature count exceeds the number of training samples it is $\nu$, not the nominal 2048, that should be weighed against the 750 training samples.

\subsection{Selecting $\lambda$}
Training error decreases monotonically as $\lambda\to0$ and so cannot select the penalty. We minimize instead the cross-validated error
\begin{equation}
	\lambda^{\star} = \arg\min_{\lambda}\ \frac{1}{K}\sum_{k=1}^{K}\text{NMSE}\Big(\mathbf{y}^{(k)},\ Z^{(k)}\boldsymbol{\beta}^{(-k)}(\lambda)\Big) ,
	\label{eq:68}
\end{equation}
where $\boldsymbol{\beta}^{(-k)}$ is fitted with fold $k$ withheld. The $K = 5$ folds are contiguous blocks, never shuffled as the samples form a time series, so a shuffled partition would place symbol $n-1$ in training and symbol $n$ in validation, and the near-identity of those states would make every $\lambda$ appear acceptable. The search is logarithmic, scanning $10^{-8}$ to $10^{6}$ at six points per decade, extending either boundary by three decades whenever the minimum lands on it, and finally refining the bracketing interval at 36 points per decade. A minimum at the edge of the grid may be truncated rather than located and could move if the grid were widened. Extension stops when three further decades improve the error by less than 0.1\%, indicating a flat curve rather than a distant optimum. Every value below lies strictly inside the searched range, with the error rising on both sides.

\subsection{Results}
\begin{table}[h]
	\centering
	\caption{Readout performance at the published regularization strength and at the cross-validated optimum.}
	\alttext{Two blocks across the four operating regimes. The first reports the effective degrees of freedom, test NMSE and train-to-test ratio at the published penalty; the second reports the cross-validated optimum, its effective degrees of freedom, test NMSE, train-to-test ratio, and the resulting improvement factor.}
	\label{tab:regularization_comparison}
	\begin{tabular}{lcccc}
		\hline
		& \textbf{Sub-threshold}
		& \textbf{Parametric resonance}
		& \textbf{Coherent comb}
		& \textbf{Chaotic comb} \\
		\hline

		\multicolumn{5}{l}{\textbf{At the published $\lambda = 10^{-3}$}} \\

		$\nu$ (of active features)
		& 4.1 / 1024
		& 63.7 / 2048
		& \textbf{703} / 2048
		& \textbf{700} / 2048 \\

		Test NMSE
		& $2.08\times10^{-2}$
		& $1.33\times10^{-3}$
		& $6.67\times10^{-2}$
		& 7.16 \\

		Train / test ratio
		& 1.2
		& 3.1
		& $\sim10^{3}$
		& $\sim10^{3}$ \\
		\hline
		\multicolumn{5}{l}{\textbf{At the cross-validated optimum}} \\

		$\lambda^{\star}$
		& $3.16\times10^{-7}$
		& 4.08
		& 21.5
		& 215 \\

		$\nu$
		& 12.4
		& 30.4
		& 225.5
		& 164.5 \\

		Test NMSE
		& $1.93\times10^{-2}$
		& $\mathbf{4.51\times10^{-4}}$
		& $1.28\times10^{-2}$
		& $6.36\times10^{-2}$ \\

		Train / test ratio
		& 1.2
		& 0.9
		& 2.3
		& 1.7 \\

		Improvement
		& 1.1$\times$
		& 3.0$\times$
		& 5.2$\times$
		& \textbf{113$\times$} \\

		\hline
	\end{tabular}
\end{table}
\begin{figure}[h]
	\centering
	\includegraphics[width=1\columnwidth]{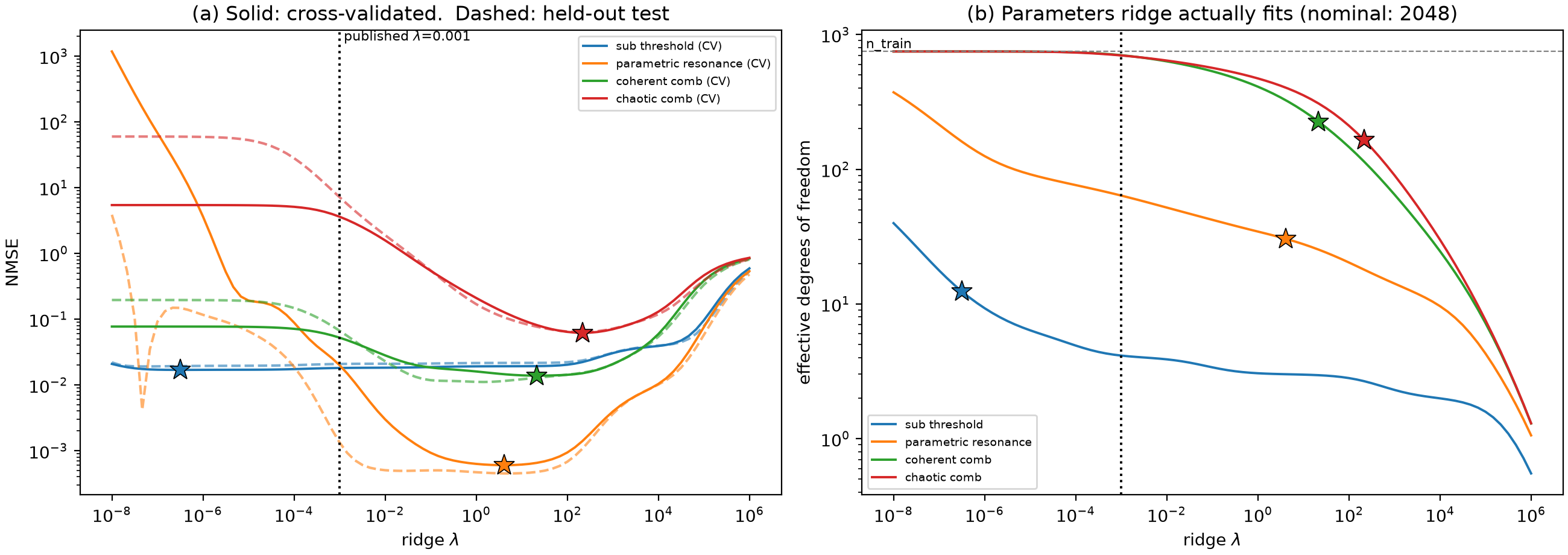}
	\caption{(a) Cross-validated (solid) and held-out test (dashed) error against $\lambda$; stars mark $\lambda^{\star}$, the dotted line the published value. (b) Effective degrees of freedom, Eq.~\ref{eq:67}.}
	\alttext{Two panels plotted against the ridge penalty on a logarithmic axis. (a) NMSE for the four regimes, with solid cross-validated curves, dashed held-out test curves, stars at each selected optimum, and a dotted vertical line at the published penalty of $10^{-3}$. Both comb curves are high and flat to the left of that line and reach shallow minima well to its right, while parametric resonance has a deep minimum near 4. (b) Effective degrees of freedom, with a dashed horizontal line at the 750 training samples. At the published penalty both comb curves lie close to that line, whereas parametric resonance is an order of magnitude below it.}
	\label{fig:penalty-sel}
\end{figure}

The effect depends on the regime. At parametric resonance, a fixed penalty of $10^{-3}$ is reasonable: by Eq.~\ref{eq:67}, the fit uses $\nu = 63.7$ directions rather than 2048 for 750 samples, with a train/test ratio of 3.1. At the comb points, this is no longer the case: $\nu \approx 700$ for 750 samples, the ratio reaches $10^{3}$, and the chaotic comb gives a test NMSE above unity, which is worse than predicting the training mean. Cross-validation reduces this error by a factor of 113. The difference comes from the flatter singular spectrum in the comb regimes. More singular values lie close to a given $\lambda$, so Eq.~\ref{eq:67} leaves more directions in the fit. These additional weak directions make the comb regimes harder to regularize.

The ordering is unchanged. At each regime's own optimum, the coherent comb still trails parametric resonance by 28$\times$. The gap becomes clearer once the comb regimes are evaluated with their appropriate penalties rather than in an overfitted condition. The penalty is well defined where the reservoir computes: the range over which the cross-validated error remains within 1\% of its minimum spans 0.50 decades at parametric resonance and the coherent comb, and 0.33 decades at the chaotic comb. Sub-threshold is different, with the error remaining within 1\% over 2.33 decades and within 10\% over 6.5 decades. Thus $\lambda$ is poorly determined there, and the published value increases the error by only 1.08$\times$.

Selecting $\lambda$ independently at 20 points spread across each regime gives an interquartile spread of 26.5$\times$ within parametric resonance alone, against a 1\% plateau of 3.2$\times$: the optima are more widely separated than any single choice can cover. Repeating the selection at one operating point over 100 startup realizations isolates the contribution of the startup state:
\begin{table}[h]
	\centering
	\caption{Distribution of cross-validated optimal regularization strengths across regimes.}
	\alttext{Median cross-validated optimal penalty, interquartile ratio and full range across 100 startup realizations, for each of the four operating regimes.}
	\label{tab:lambda_distribution}
	\begin{tabular}{lccc}
		\hline
		& \textbf{Median $\lambda^{\star}$}
		& \textbf{Interquartile ratio}
		& \textbf{Full range} \\
		\hline
		Sub-threshold
		& $3.16\times10^{-7}$
		& $1.00\times$
		& $1\times$ \\

		Parametric resonance
		& 4.084
		& $1.00\times$
		& $1\times$ \\

		Coherent comb
		& 2.97
		& $129\times$
		& $464\times$ \\

		Chaotic comb
		& 261
		& $1.29\times$
		& $2.61\times$ \\
		\hline
	\end{tabular}
\end{table}
At sub-threshold and parametric resonance, the startup protocol has no effect: all 100 realizations return the same values, as expected for contracting dynamics. Across the full range of realizations the chaotic comb varies by only $2.61\times$, with mixing giving a unique invariant measure whose statistics are reproduced even though the individual trajectories differ. The coherent comb, which is marginal by Eq.~\ref{eq:55} and therefore neither contracting nor mixing, varies by $464\times$. Its median of $2.97$ therefore differs from the $21.5$ of Table~\ref{tab:regularization_comparison}, which reports a single realization that happened to fall in the upper cluster. The two are consistent draws from the same bimodal distribution rather than conflicting measurements. Its distribution separates into two clusters, split almost evenly at $\lambda\approx1.7$ (49 versus 51), with a separation of 3.5 times the combined standard deviation of the two clusters. This is not the symmetry implied by Eq.~\ref{eq:46}, which produces identical feature matrices, but instead points to two distinct states retaining 510 and 226 effective directions, respectively.

The parameter sweeps were repeated under three policies: the fixed penalty, a per-regime penalty, and cross-validation at every grid point.
\begin{figure}
	\centering
	\includegraphics[width=1\columnwidth]{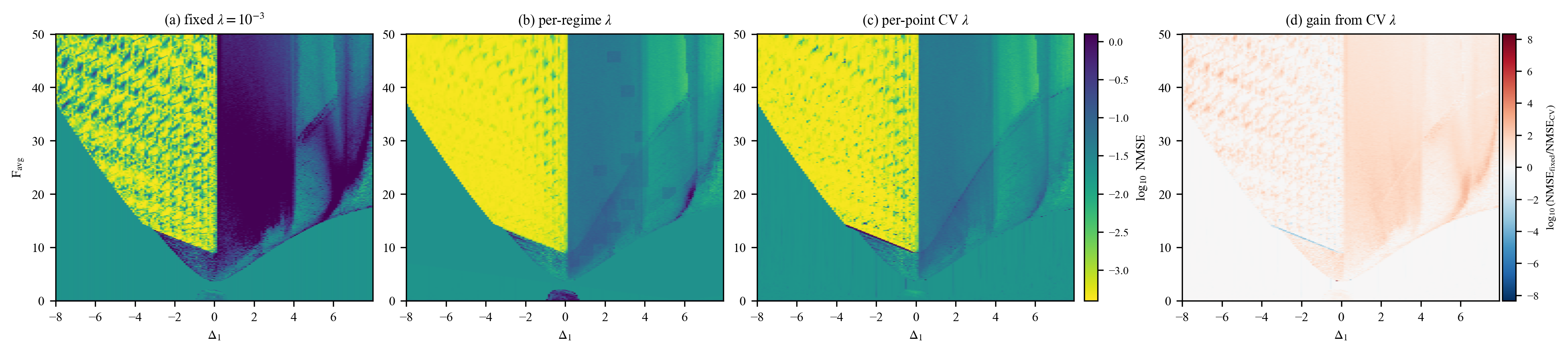}
	\caption{$\log_{10}$ NMSE over the $(\text{F}_\text{avg},\Delta_1)$ plane at (a) fixed $\lambda = 10^{-3}$, (b) per-regime $\lambda$, (c) per-point cross-validated $\lambda$, on a common color scale. (d) The gain from cross-validation over the fixed penalty.}
	\alttext{Four maps over the plane of average drive amplitude against detuning. Panels (a), (b) and (c) show the base-ten logarithm of NMSE under a fixed penalty, a per-regime penalty and a per-point cross-validated penalty, on a shared colour scale. The comb region at positive detuning is dark, close to NMSE of one, under the fixed penalty and becomes uniformly lighter under the two corrected policies, while the speckle in the parametric-resonance region of panel (a) largely disappears. Panel (d) shows the ratio between the fixed and cross-validated errors, red across most of the plane, indicating improvement almost everywhere.}
\end{figure}
\begin{table}[h]
	\centering
	\caption{Relative trajectory error over 400 symbols, measured against the tightest setting.}
	\alttext{Relative trajectory error over 400 symbols at three relative tolerances, together with the number of right-hand-side evaluations per symbol, for each of the four operating regimes.}
	\label{tab:integration_convergence}
	\begin{tabular}{lcccc}
		\hline
		\textbf{Relative tolerance}
		& \textbf{Sub-threshold}
		& \textbf{Parametric resonance}
		& \textbf{Coherent comb}
		& \textbf{Chaotic comb} \\
		\hline
		$10^{-6}$
		& $6.7\times10^{-7}$
		& $2.1\times10^{-6}$
		& $1.2\times10^{-3}$
		& 1.19 \\

		$10^{-8}$ (used)
		& $7.0\times10^{-9}$
		& $2.1\times10^{-8}$
		& $7.0\times10^{-6}$
		& 1.20 \\

		$10^{-10}$
		& $7.2\times10^{-11}$
		& $2.3\times10^{-10}$
		& $4.9\times10^{-8}$
		& 1.20 \\

		\textbf{RHS evaluations per symbol}
		& \textbf{508}
		& \textbf{1765}
		& \textbf{5295}
		& \textbf{3502} \\
		\hline
	\end{tabular}
\end{table}
Cross-validation improves 91.2\% of grid points, with a median gain of 0.41 decades. The improvement is concentrated in the comb region, which at the fixed penalty lies close to $\text{NMSE} = 1$ throughout and recovers to ${\sim}10^{-2}$ once the penalty is selected; the speckle in the parametric-resonance region of panel (a) is readout variance and largely disappears under either corrected policy. The ordering reported in the main text survives: under per-point cross-validation, parametric resonance outperforms the coherent comb by $8.0\times$ and the chaotic comb by $25.5\times$.
\begin{figure}[h]
	\centering
	\includegraphics[width=1\columnwidth]{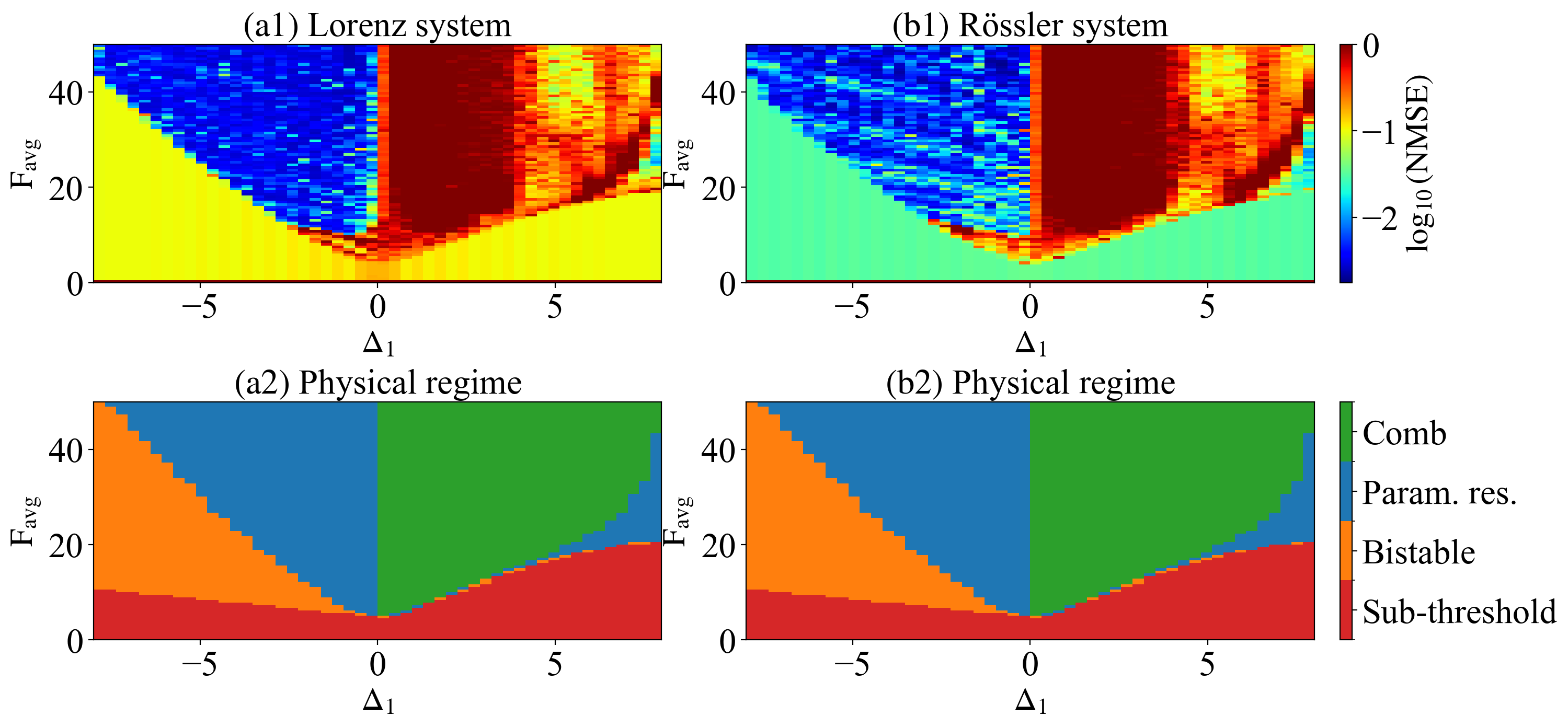}
	\caption{(a1, b1) Normalized mean squared error (NMSE, log scale) as a function of average drive amplitude $\text{F}_\text{avg}$ and detuning $\Delta_1$, with the corresponding dynamical regime maps in (a2, b2) showing sub-threshold (ST), parametric resonance (PR), and frequency comb (FC) regions for Lorenz time-series and Rössler time-series. The NMSE pattern closely mirrors the NMSE plot of Mackey--Glass in the main text. Note: $\gamma_{21}=1.0$, $\kappa=-9.0$ and input data rate = 2454 samples/second.}
	\alttext{Four panels. (a1) and (b1) show the base-ten logarithm of NMSE over the plane of average drive amplitude against detuning for the Lorenz and Rössler benchmarks; both show a blue low-error wedge at negative detuning and dark red high-error regions at positive detuning. (a2) and (b2) show the corresponding four-level regime maps, with sub-threshold red, bistable orange, parametric resonance blue and comb green. The low-error wedge coincides with the parametric-resonance region in both benchmarks.}
	\label{fig:lorenz-rossler-sweep}
\end{figure}
Across the points reaching $\text{NMSE} < 10^{-3}$, $\lambda^{\star}$ spans $8.3\times10^{-5}$ to $29.7$ with an interquartile range of $14.7\times$, and the value $10^{-3}$ lies at the $0.5$th percentile of that distribution. Per-point cross-validated $\lambda$ is therefore adopted.

\section{Numerical integration}
\label{sec:num-int}
Eqs.~\ref{eq:35}--\ref{eq:36} are integrated as a real four-component system using an adaptive Runge--Kutta method of order 5(4), with relative tolerance $10^{-8}$ and absolute tolerance $10^{-10}$. The integrator is called once per symbol at constant drive, with the terminal state carried forward. Since the stepping is adaptive, there is no fixed step count, so the relevant measure of computational cost is the number of right-hand-side evaluations per symbol.

Every non-chaotic point converges cleanly, with roughly two orders of magnitude improvement in trajectory error for every two orders of magnitude tightening in tolerance. The chaotic comb does not converge pointwise, as expected for a chaotic system: an initial discrepancy grows as
\begin{equation}
	\delta(\tau) \simeq \delta_0\, e^{\lambda_\text{max}\tau} .
	\label{eq:69}
\end{equation}
With $\lambda_\text{max} = +0.563$ per $\tau$ over the $400\times5\tau = 2000\,\tau$ used in this comparison, the amplification factor is $e^{1126}$. Thus, no practical tolerance can keep two trajectories together. The relative error settles at approximately 1.2, which is the expected separation between two uncorrelated points on the same attractor. This is a property of the chaotic regime, not a failure of the integrator. The chaotic comb should therefore be described using statistical quantities such as $\lambda_\text{max}$, spectral entropy, and prediction error over a long record, rather than by comparing individual trajectories. The same result is obtained independently in the startup tests.

For stiffness, an explicit eighth-order method and two implicit methods agree with the reference to $2\times10^{-8}$ at the same tolerances. However, the implicit Radau method costs between 2.28 and 3.93 times more than the explicit method across the four regimes. The relevant diagnostic is computational cost rather than agreement: for a stiff system, an implicit method should allow much larger steps because its step size is not limited by the fastest timescale. That does not occur here. The system is therefore not stiff in any of the regimes examined, including the chaotic comb, and the adaptive explicit method is appropriate.

\begin{figure}[htbp]
	\centering
	\includegraphics[width=1\columnwidth]{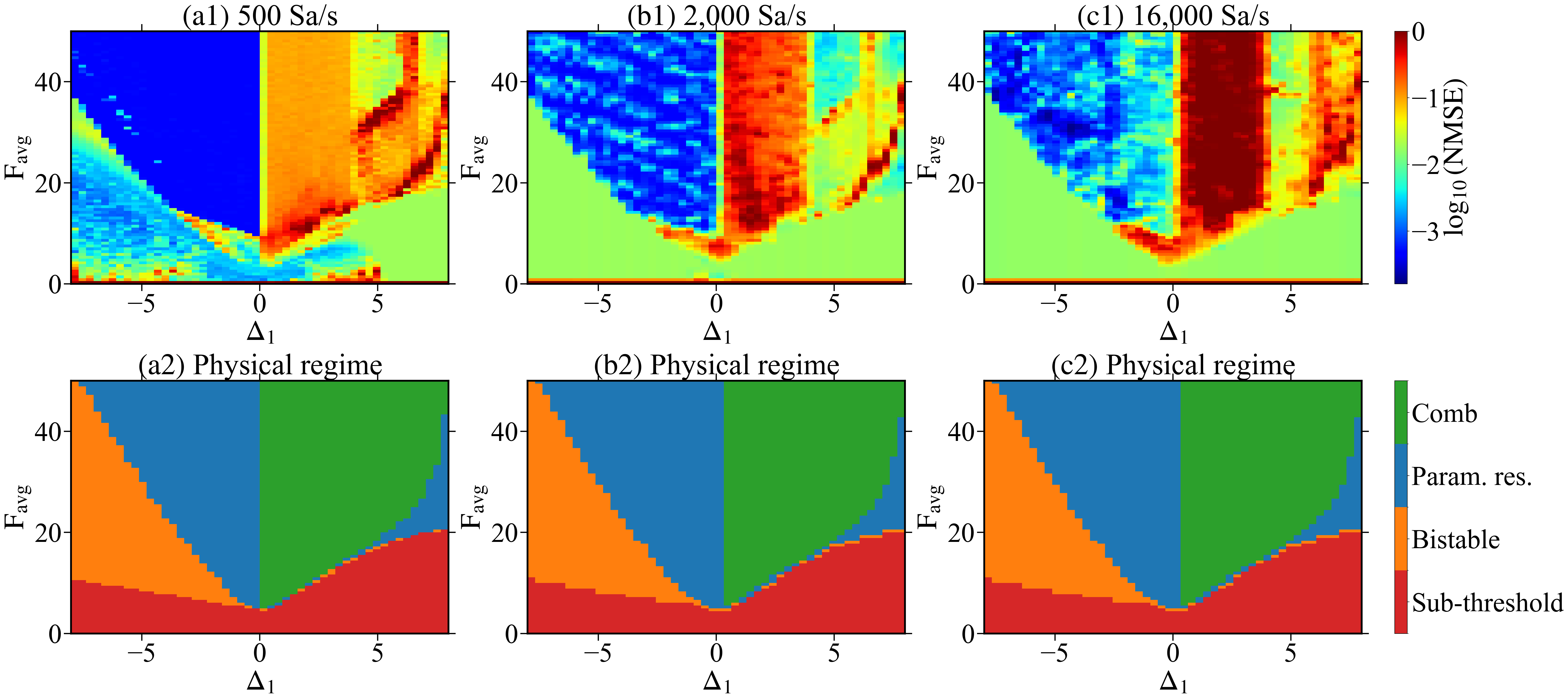}
	\caption{(a1--c1) Dependence of prediction performance on input data rate, with the corresponding dynamical regime maps in (a2--c2). NMSE (log scale) is shown over the $\left(\text{F}_\text{avg}, \Delta_1 \right)$ parameter
		space for different data rates. Optimal performance is observed at intermediate data rates, where the input timescale matches the intrinsic dynamical response of the system. Note: $\Delta \text{F} = 0.0,\ \gamma_{21}=1.0,\ \kappa=-9.0$.}
	\alttext{Six panels in two rows. The top row shows the base-ten logarithm of NMSE over the plane of average drive amplitude against detuning at 500, 2000 and 16,000 samples per second; the blue low-error region inside parametric resonance is largest at the lowest rate and contracts as the rate rises, while the comb region at positive detuning becomes uniformly dark red. The bottom row shows the corresponding four-level regime maps, which are essentially unchanged across the three rates.}
	\label{fig:data-rate-sweep}
\end{figure}
\begin{figure}[htbp]
	\centering
	\includegraphics[width=1\columnwidth]{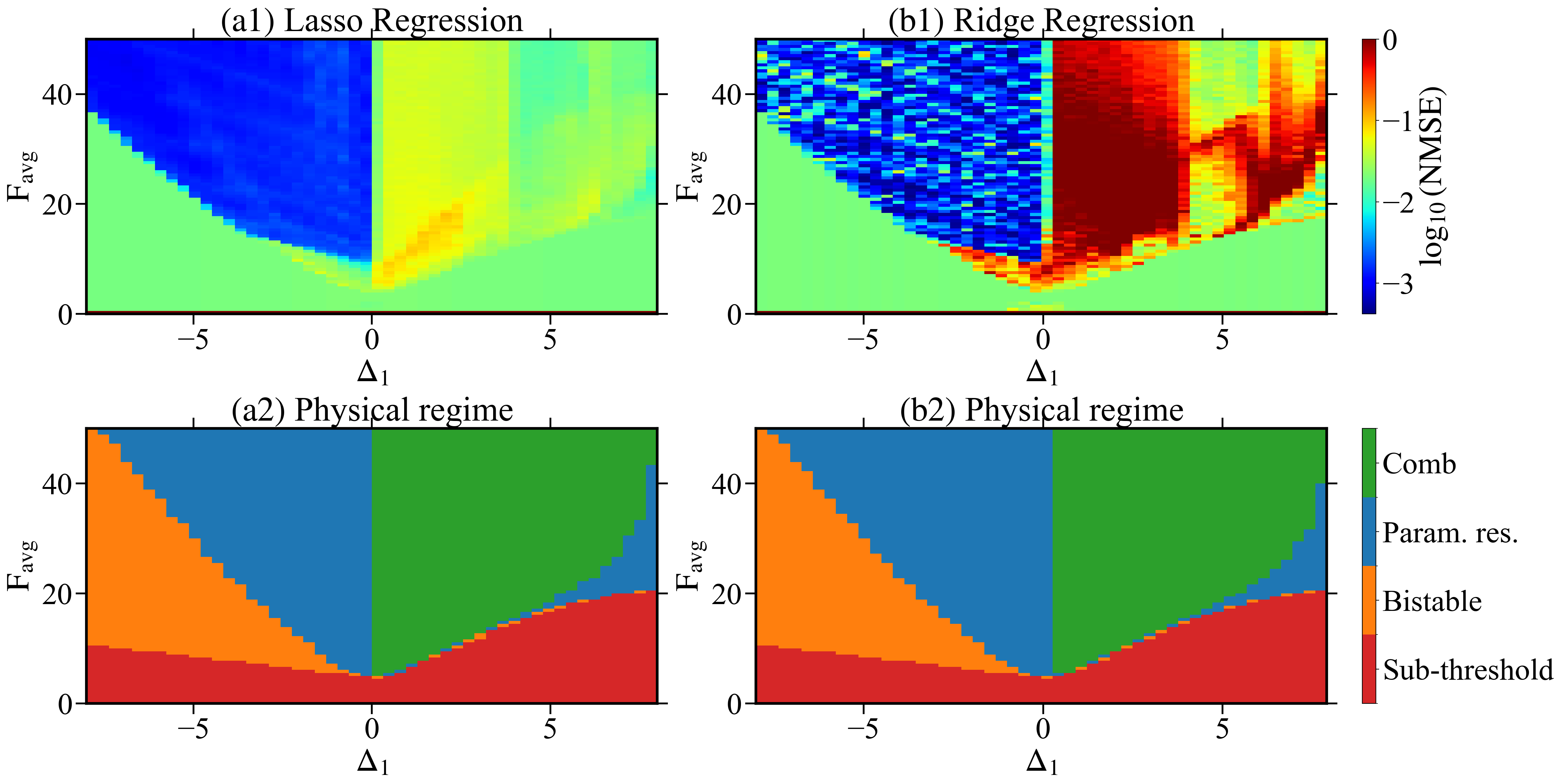}
	\caption{Comparison of prediction performance obtained using LASSO and ridge regression readouts for the Mackey--Glass time series. (a1) NMSE (log scale) obtained using LASSO regression over the $\left(\text{F}_\text{avg}, \Delta_1 \right)$ parameter space. (b1) Corresponding NMSE obtained using ridge regression. LASSO achieves improved prediction performance across both the parametric resonance and frequency comb regimes while utilizing a reduced set of active features. The results indicate that sparse readouts can effectively capture the essential nonlinear dynamics of the reservoir. Note: $\Delta \text{F}=0.0,\ \gamma_{21}=1.0,\ \kappa=-9.0$ and input data rate = 2454 samples/second.}
	\alttext{Four panels. (a1) shows the base-ten logarithm of NMSE over the plane of average drive amplitude against detuning for a LASSO readout and (b1) the same for a ridge readout, on a shared colour scale. The LASSO map is smoother and reaches lower error across both the parametric-resonance wedge and the comb region, whereas the ridge map is heavily speckled at negative detuning and largely dark red at positive detuning. (a2) and (b2) show the corresponding four-level regime maps, which are the same in both cases.}
	\label{fig:mackey-glass-lasso}
\end{figure}
\section{Additional parameter sweeps for Lorenz and Rössler systems}
\label{sec:extra-sweeps}
To verify the generality of the observed computational behavior, we extend the parameter-space analysis presented in the main text to additional chaotic benchmarks, namely the Rössler and Lorenz systems.
Fig.~\ref{fig:lorenz-rossler-sweep} (a1) and (b1) show the normalized mean squared error (NMSE) as a function of the average drive amplitude $\text{F}_\text{avg}$ and detuning $\Delta_1$, along with the corresponding dynamical regime maps: Fig.~\ref{fig:lorenz-rossler-sweep} (a2) and (b2). The same parameter ranges and numerical procedures as described in the main text are used.

Across both systems, the lowest NMSE values are consistently localized within the parametric resonance regime. This confirms that the alignment between optimal computational performance and the underlying bifurcation structure is not specific to a particular input signal, but rather a general feature of the parametrically driven reservoir.

In contrast, the frequency comb regime exhibits heterogeneous performance with regions of moderate prediction performance interspersed with high-error zones. This behavior is consistent with the loss of phase coherence observed in the chaotic comb regime, which degrades the structured nonlinear transformations required for effective prediction.
\section{Dependence on input data rate}
\label{sec:data-rate}
We further investigate the influence of input data rate on reservoir performance. Figure~\ref{fig:data-rate-sweep} shows the NMSE over the $\left(\text{F}_\text{avg}, \Delta_1 \right)$ parameter space for multiple input data rates spanning low, intermediate, and high regimes.

At low data rates as in Fig.~\ref{fig:data-rate-sweep} (a1), the symbol duration is long compared to the intrinsic relaxation timescale of the system. As a result, the reservoir dynamics settle between the successive inputs, leading to reduced memory and poor prediction performance.

At intermediate data rates as shown in Fig.~\ref{fig:data-rate-sweep} (b1), a well-defined low-NMSE region emerges within the parametric resonance regime. In this regime, residual dynamical activity overlaps successive inputs, enabling both memory retention and nonlinear transformation, which are essential for reservoir computing.

At high data rates, Fig.~\ref{fig:data-rate-sweep} (c1), the system is unable to fully respond within each symbol period. This leads to degraded performance and a contraction of the low-error region. These results indicate that optimal computational performance is achieved when the input timescale is matched to the intrinsic dynamical response of the system.
\section{LASSO regression based readout}
\label{sec:lasso}
In the main text, the reservoir readout is trained using ridge regression. Here, we replace the ridge regression with a LASSO (Least Absolute Shrinkage and Selection Operator) readout to investigate the effect of sparsity on computational performance. LASSO introduces an $L_1$ regularization term, promoting sparse solutions in the readout weights. This effectively performs feature selection by retaining only the most relevant components of the high-dimensional reservoir state.
\begin{figure}[htbp]
	\centering
	\includegraphics[width=1.0\columnwidth]{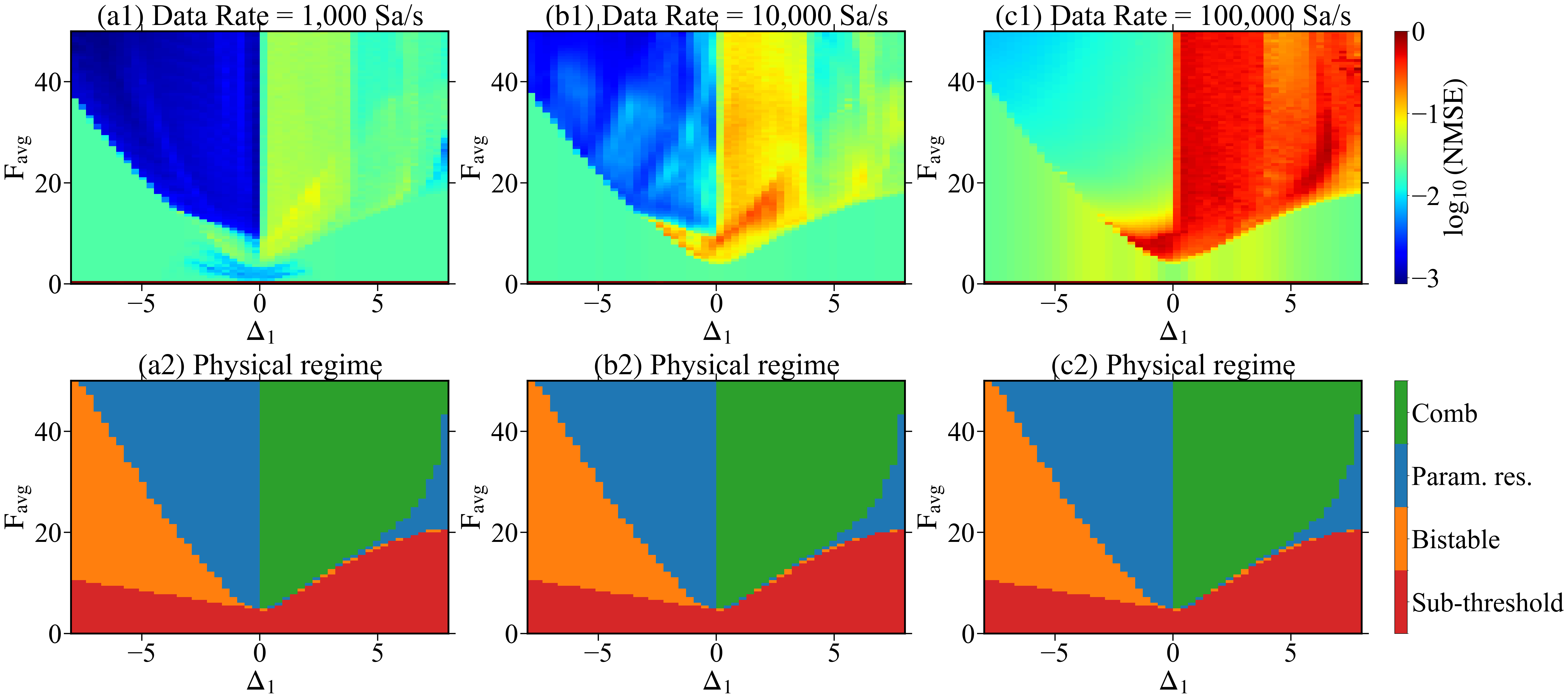}
	\caption{Dependence of prediction performance on input data rate using LASSO regression. NMSE (log scale) is shown over the $\left(\text{F}_\text{avg}, \Delta_1 \right)$ parameter space for different data rates. Similar to ridge regression, optimal performance is observed at intermediate data rates where the input timescale matches the intrinsic dynamical response of the system. The use of LASSO results in a reduced effective dimensionality of the reservoir while preserving the overall computational trends across dynamical regimes. Note: $\Delta \text{F}=0.0,\ \gamma_{21}=1.0,\ \kappa=-9.0$.}
	\alttext{Six panels in two rows. The top row shows the base-ten logarithm of NMSE over the plane of average drive amplitude against detuning for a LASSO readout at 1000, 10,000 and 100,000 samples per second. The low-error region is deepest at the lowest rate and degrades progressively, becoming largely dark red at the highest. The bottom row shows the corresponding four-level regime maps, unchanged across the three rates.}
	\label{fig:mackey-glass-lasso-datarate-sweep}
\end{figure}
Fig.~\ref{fig:mackey-glass-lasso} (a1) shows the prediction performance obtained using LASSO regression on the Mackey--Glass time series and Fig.~\ref{fig:mackey-glass-lasso} (b1) shows the performance for ridge regression. LASSO achieves better prediction performance compared to ridge regression in both the parametric resonance and frequency comb regions. The number of active features is significantly reduced, indicating that only a subset of the reservoir states contributes meaningfully to the prediction task.
These results suggest that the computational capability of the reservoir is not uniformly distributed across all features, and that sparse readouts can capture the essential dynamics with reduced complexity.
\section{Data rate dependence with LASSO}
\label{sec:lasso-data-rate}
To further examine the interplay between sparsity and temporal processing, we evaluate the performance of the LASSO readout across different input data rates.
Fig.~\ref{fig:mackey-glass-lasso-datarate-sweep} shows the NMSE over the $\left(\text{F}_\text{avg}, \Delta_1 \right)$ parameter space for multiple data rates using LASSO. The overall trends remain consistent with those observed using ridge regression:
\begin{enumerate}
	\item Optimal performance occurs at intermediate data rates within the parametric resonance regime.
	\item Low data rates lead to loss of memory, while high data rates limit the system's ability to respond dynamically.
\end{enumerate}
The use of LASSO regression results in a reduced effective dimensionality of the reservoir. In some cases, this improves robustness at higher data rates by suppressing noisy or redundant features. At the same time, excessive sparsity can limit performance in regimes where richer feature representations are required.

\makeatletter
\patchcmd{\endthebibliography}{\label{LastBibItem}}{\label{LastBibItemSupp}}{}{}
\def\NAT@bibsetnum#1{\setlength{\topsep}{\z@}\NATx@bibsetnum{\ref{LastBibItemSupp}}}
\makeatother

\end{document}